\documentclass[11pt]{article}

\usepackage[final]{acl}
\usepackage{amsmath}

\usepackage{times}
\usepackage{latexsym}

\usepackage[T1]{fontenc}

\usepackage[utf8]{inputenc}
\usepackage{gensymb}

\usepackage{microtype}

\usepackage{inconsolata}

\usepackage{graphicx}
\usepackage{booktabs}
\usepackage{multirow}
\usepackage{tabularx}
\usepackage{verbatim}

\usepackage[utf8]{inputenc}
\usepackage[T1]{fontenc}

\usepackage{mdframed}
\usepackage[backgroundcolor=blue!20, bordercolor=blue, linecolor=blue!50]{todonotes}
\usepackage{hyperref}

\newcommand{\sectioncolor}{violet}

\title{GeoAgent: Evaluating VLM Geolocalization Through Embodied Navigation}

\author{Arka Mukherjee \\
  KIIT Bhubaneswar \\
  \texttt{arka.mukherjee078@gmail.com} \\\And
  Soham Roy \\
  KIIT Bhubaneswar \\
  \texttt{sohamroy.dev@gmail.com} \\\AND
  Kartikeya Trivedi \\
  KIIT Bhubaneswar \\
  \texttt{kartikeya.trivedi@gmail.com} \\\And
  Shreya Ghosh \\
  IIT Bhubaneswar \\
  \texttt{shreya@iitbbs.ac.in} \\
  }

\begin{document}
\maketitle
\begin{abstract}
Modern Vision-Language Models (VLMs) perform well above the human baseline in image geolocalization, a task critically important in disaster response, OSINT verification, and location privacy. However, most efforts to study AI behavior on the task remain limited to static image-based retrieval, classification, and predictions. We argue that faithful recreation of the task should involve embodied navigation, where a multimodal agent autonomously explores its surroundings to gather observations before submitting a prediction. To this end, we introduce \textbf{GeoAgent}, an agentic environment-based benchmark that requires agents to navigate Street View environments to refine their geolocalization through sequential reasoning. Our analysis shows that modern VLMs struggle to discern regional patterns while succeeding at country- and continent-level predictions. When compared to static image-based baselines, agentic navigation significantly improves accuracy across established metrics. We also note severe bias in a developed/developing region context across frontier model architectures and poor self-improvement capabilities given incorrect priors. Overall, our work establishes the challenges of embodied navigation and geospatial reasoning. We publicly release our code and the GeoAgent environment: \url{https://geoagent-benchmark.github.io}
\end{abstract}

\section{Introduction}

Geolocalization from visual inputs has emerged as a challenging spatial reasoning task for modern AI systems. Among various tests, the popular game GeoGuessr\footnote{\url{https://www.geoguessr.com}} attracts the most attention~\cite{haas2024pigeonpredictingimagegeolocations,cheng2025geoguessmultimodalreasoningbased, talreja2026georcbenchmarkgeolocationreasoning, Wazzan2024ComparingTA, Jay2025EvaluatingPG,suresh2018deepgeophotolocalizationdeep}, which requires accurate geospatial predictions within a limited Street View-like navigable environment. Early AI approaches to this game framed GeoGuessr as an image classification task~\citep{haas2024pigeonpredictingimagegeolocations}. More recently, Vision-Language Models (VLMs) have demonstrated remarkable zero-shot performance on geolocalization 
tasks~\citep{cheng2025geoguessmultimodalreasoningbased, yi2025geolocsftefficientvisualgeolocation, zhang2025navig, li2025recognitionreasoningreinforcingimage}, owing to their web-scale generalizability in interpreting visual cues such as signage, architecture, and vegetation.

\begin{figure}
    \centering
    \includegraphics[width=0.85\columnwidth]{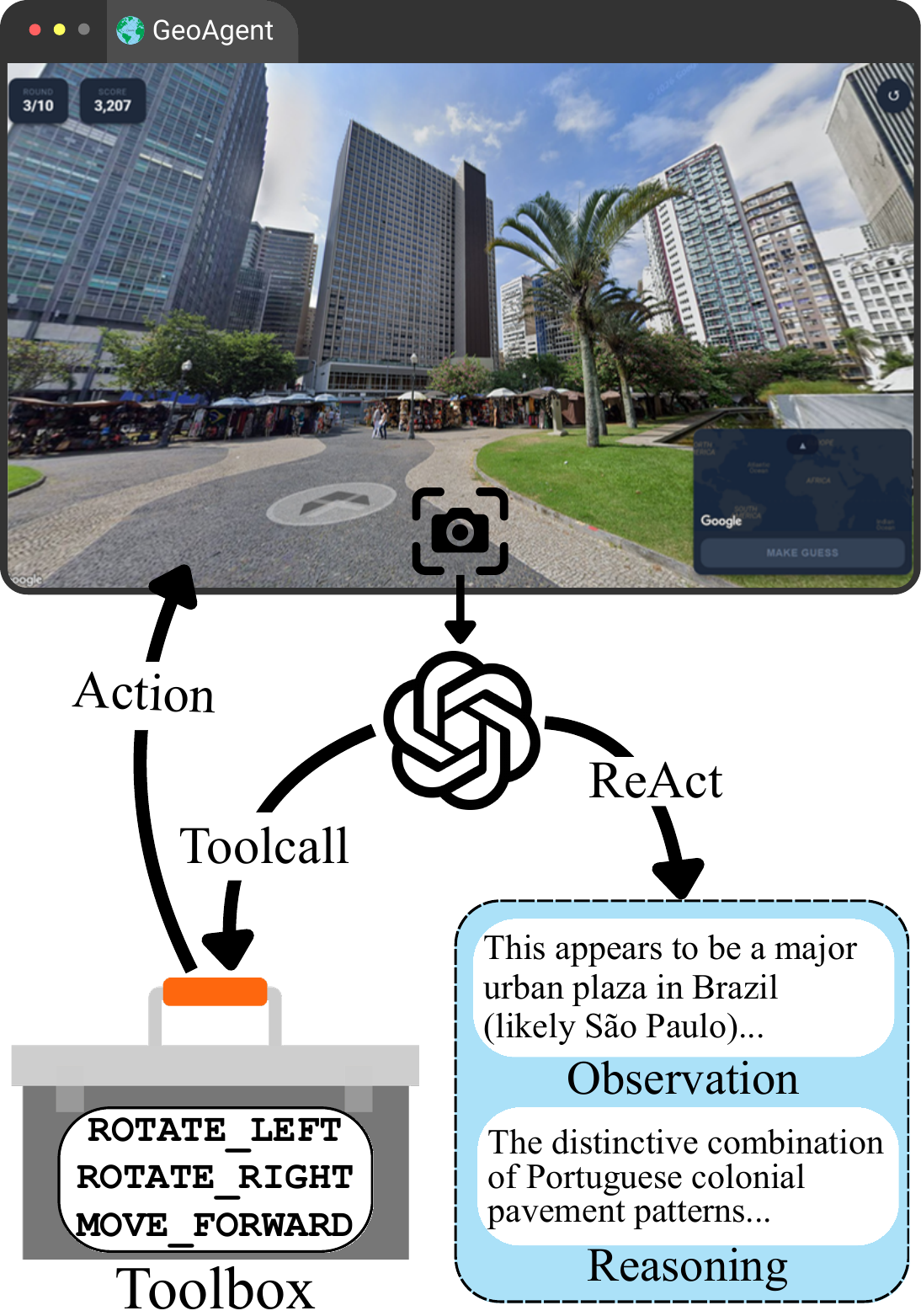}
    \caption{An overview of the GeoAgent framework, with an example VLM (GPT-5-Mini) interacting. Each agent is provided with a toolbox containing nine actions to explore the environment and submit a guess through iteratively refining their geolocalization prediction.}
    \label{fig:geoagent_summary}
\end{figure}

Beyond the game, visual geolocalization finds use in many real-world workflows, such as disaster response, where damage-assessment pipelines must localize ground-level photographs of affected areas when GPS metadata are stripped, spoofed, or never recorded~\cite{Gupta_2019_CVPR_Workshops}. Further, it also finds usage in OSINT verification, fact-checking~\cite{song2025geocomp}, autonomous navigation, and robotics. Understanding the limits of this capability is important to prevent misuse~\citep{huang-etal-2025-ai, li2025pixelsplacessystematicbenchmark}.

Most existing evaluations assess models on single, static images, which fail to capture how humans actually tackle geolocalization through active exploration of their environment. \citet{vyas2022gama} provides richer contextual clues using video clips, but remains a passive observer. Passive inference traps the model in a fixed trajectory: critical identifying features cannot be captured by a constrained set of static images, and the most informative view for a given location may simply not be among them, leaving no room to seek disambiguating evidence. Embodied navigation resolves this bottleneck by letting the agent reason and execute navigational actions, such as movement and rotation, to maximize information gain. This mirrors how human players approach geolocalization and represents an unexplored frontier for evaluating spatial reasoning in VLMs. Evaluating VLMs only under passive inference therefore tests recognition, not reasoning under uncertainty, which is the distinction we achieve with GeoAgent.

Modern advances in tool calling~\cite{patil2025the,yao2023reactsynergizingreasoningacting,erdogan2024tinyagent,Ehtesham2025ASO,Singh2025AgenticRA} and agentic web navigation~\cite{wu2025webwalker,deng2023mind2web,zhou2023webarena} have made simulating web-based workflows possible. Leveraging these capabilities, we introduce \textbf{GeoAgent}, both an open-source, modular, and reusable agentic environment and a benchmark that evaluates VLMs through embodied navigation in fully explorable Street View environments (Figure~\ref{fig:geoagent_summary}). The task involves two distinct steps: exploration actions with tool calls and geospatial reasoning for the final location guess. This enables us to study the following research questions that static evaluation does not capture:

\begin{itemize}
    \setlength{\itemsep}{0pt}
    \setlength{\parskip}{0pt}
    \item \textbf{RQ1:} Do VLMs improve geolocalization accuracy through active exploration, and does improvement depend on the correctness of initial hypotheses?
    \item \textbf{RQ2:} Do frontier VLMs exhibit systematic geographic bias between developed and developing regions?
    \item \textbf{RQ3:} How do exploration strategies differ across VLM architectures, and does action efficiency correlate with task performance?
\end{itemize}

\begin{figure}
    \centering
    \includegraphics[width=\columnwidth]{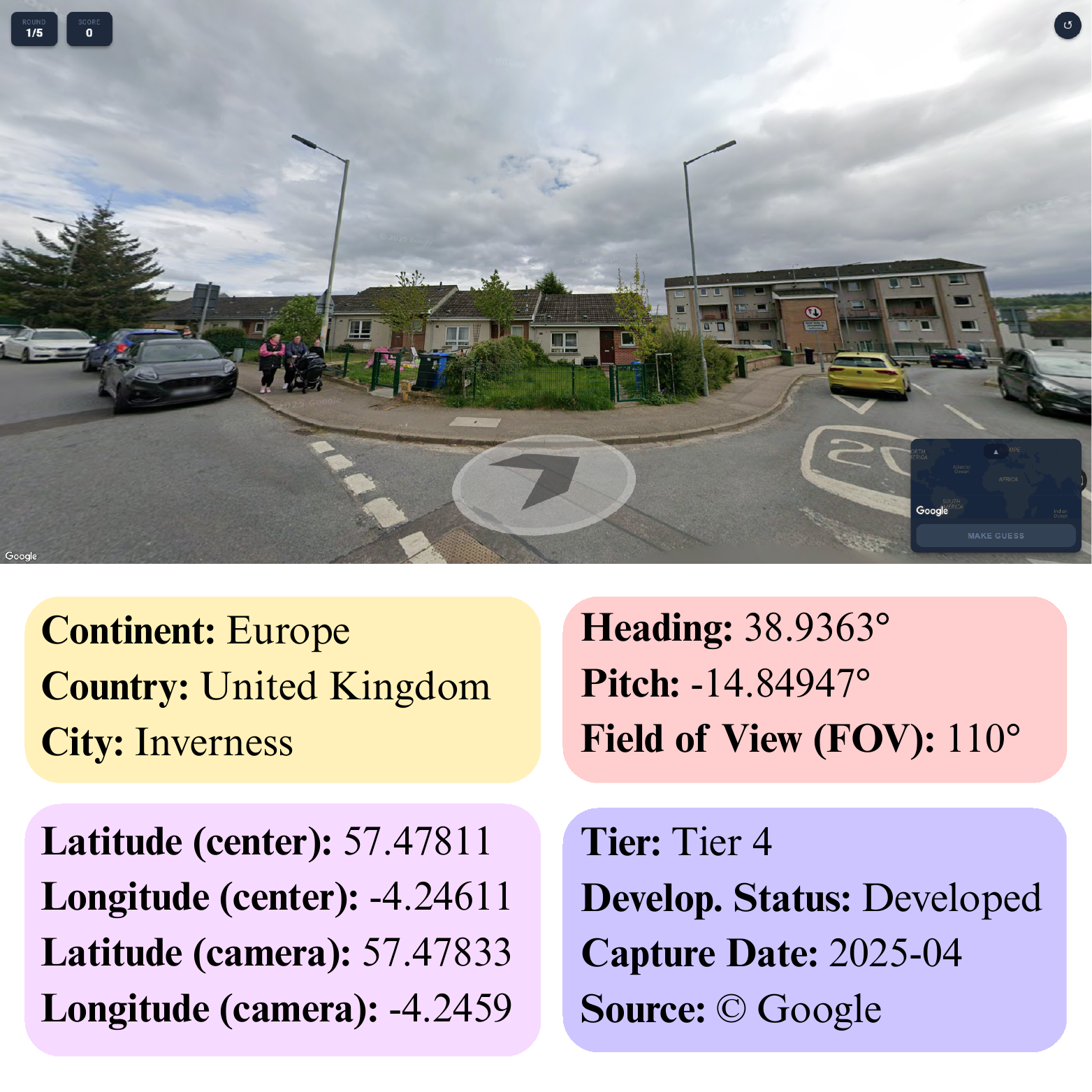}
    \caption{A screenshot of the navigable GeoAgent interface with the collected data.}
    \label{fig:example_data}
\end{figure}

Our benchmark comprises 1,200 fully navigable Street View locations across 100 cities spanning developed and developing regions worldwide, with difficulty stratified by recognizability tiers, as shown in Figure~\ref{fig:example_data}. We conduct comprehensive agentic behavioral evaluations of five frontier VLMs plus Geo-R1 7B and compare them to static multiview baselines, revealing that models demonstrate limited self-correction capabilities when initial hypotheses are incorrect. Our analyses also reveal substantial bias among developed and developing geolocations across frontier model capabilities. To the best of our knowledge, our work represents the first comprehensive study of agentic behavior under geospatial embodied navigation constraints. We plan to release our data publicly upon acceptance to promote further research.

\section{Related Work}

\begin{table*}
\centering
\footnotesize
\begin{tabular}{lcccccc}
\toprule
\textbf{Tier} & \textbf{\# Cities} & \textbf{Samples} & \textbf{Share (\%)} & \textbf{Example Cities} & \textbf{Developed} & \textbf{Developing} \\
\midrule
Tier 1 & 5  & 34  & 2.85  & New York, Tokyo      & 5  & 0  \\
Tier 2 & 10 & 86  & 7.14  & Rome, Mumbai        & 5  & 5  \\
Tier 3 & 25 & 257 & 21.42 & Bern, Bangalore     & 10 & 15 \\
Tier 4 & 25 & 343 & 28.56 & Oulu, Mombasa       & 12 & 13 \\
Tier 5 & 35 & 480 & 40.00 & Inuvik, Siem Reap   & 14 & 21 \\
\midrule
\textbf{Total} & \textbf{100} & \textbf{1,200} & \textbf{100.00} & -- & \textbf{46} & \textbf{54} \\
\bottomrule
\end{tabular}
\caption{Tier-wise composition of the final dataset (1,200 samples) with example cities and development status.}
\label{tab:tier_dataset_composition}
\end{table*}

\begin{table}[ht]
\footnotesize
\centering
\resizebox{\columnwidth}{!}{%
\begin{tabular}{ll}
\hline
\textbf{Action} & \textbf{Description} \\
\hline
ROTATE:\textless degrees\textgreater & Rotate to a specific heading (0--360°) \\
ROTATE\_LEFT & Rotate 90° left from current view \\
ROTATE\_RIGHT & Rotate 90° right from current view \\
ROTATE\_BEHIND & Rotate 180° to look behind \\
LOOK\_UP & Tilt the camera up by 20° \\
LOOK\_DOWN & Tilt the camera down by 20° \\
MOVE\_FORWARD & Move forward along the road or path \\
RETURN\_START & Return to the starting position \\
GUESS & Submit the final location guess \\
\hline
\end{tabular}%
}
\caption{Available tool calls for agents to explore the GeoAgent environment.}
\label{tab:streetview_actions}
\end{table}


\subsection{Vision-based Geolocalization}
Early foundational works treated planetary-scale geolocalization as a discrete prediction task. \textit{PIGEON} \citep{haas2024pigeonpredictingimagegeolocations} established state-of-the-art performance using semantic geocell clustering and contrastive pretraining on \textit{GeoGuessr} data. \citet{yi2025geolocsftefficientvisualgeolocation} democratized this with \textit{GeoLocSFT}, achieving competitive results via efficient supervised fine-tuning on small, high-quality datasets (${\sim}$2.7K pairs). Despite their effectiveness, these specialist models operate as black boxes on single static images, lacking interpretability.

To address this, the recent literature uses Vision-Language Models (VLMs) for explicit reasoning. \textit{GeoReasoner} \citep{10.5555/3692070.3693246} employs reasoning-tuning to extract visual cues (e.g., signage, vegetation), while \textit{GLOBE} \citep{li2025recognitionreasoningreinforcingimage} introduces a bi-objective optimization framework to assess image localizability before predicting. Concurrently, \textit{Geo-R1} \citep{xu2025geor1unlockingvlmgeospatial} utilizes reinforcement learning to instill a geospatial thinking paradigm, and \textit{GeoGuess} \citep{cheng2025geoguessmultimodalreasoningbased} establishes benchmarks for hierarchical reasoning (Country ${\rightarrow}$ City). However, evaluations like \textit{IMAGEO-Bench} \citep{li2025pixelsplacessystematicbenchmark} and \textit{FairLocator} \citep{huang-etal-2025-ai} reveal that general-purpose VLMs (GPT-4V and Gemini) exhibit geographic biases in non-Western regions.  \textit{GAMa} \citep{vyas2022gama} substitutes static images with video clips for cross-view geo-localization, yet the observer cannot act to resolve ambiguity. Crucially, all these approaches remain confined to \textit{passive} inference by analyzing fixed views, unlike GeoGuessr.

\subsection{Agentic Web Navigation}
Recently, AI evaluation has shifted toward autonomous interaction, primarily in 2D web environments. Benchmarks like \textit{WebArena} \citep{zhou2023webarena} and \textit{Mind2Web} \citep{deng2023mind2web} evaluate agents on long-horizon tasks requiring DOM manipulation. Advanced methods have since optimized this agency: \textit{Synapse} \citep{zheng2024synapse} introduces memory-augmented prompting (trajectory-as-exemplar) for complex workflows, \textit{RealWorldWebAgent} \citep{gur2024a} employs efficient planner models (HTML-T5), and \textit{WebWalker} \citep{wu2025webwalker} focuses on systematic subpage traversal. In the geospatial domain, agency remains nascent. \textit{NAVIG} \citep{zhang2025navig} introduces limited tool use (map retrieval, knowledge bases), and \textit{GeoComp} \citep{song2025geocomp} models human-like deduction chains. However, these geo-agents remain physically static, unable to navigate the environment, while web agents operate in semantic (not spatial) 2D spaces.

\textit{GeoAgent} bridges this gap by treating geolocation as \textit{embodied, interactive exploration}. Unlike web agents navigating DOM trees, we formalize navigation actions (movement, rotation) within 3D environments (Google Street View). By evaluating models' ability to actively reduce uncertainty through physical exploration, we establish a new paradigm for assessing spatial intelligence.

\section{Method}

\subsection{Task Setup \& Data Collection}
\label{sec:task-setup}

GeoAgent involves a navigable web interface that can be navigated through tool calls to explore the surrounding street signs, vegetation, and architecture for the final guess. We design a minimal interface that simulates navigation environments akin to OpenGuessr\footnote{\url{https://openguessr.com}} with Google Street View. VLMs must autonomously explore a provided environment, explain their observations, and reason through a guess at every step. 

A \textbf{round} is one full evaluation instance from a single spawn point in the dataset, played once by one model. An \textbf{action} is one complete agent turn within a round, comprising a rendered viewport screenshot, a model call that returns the structured JSON of Table~\ref{tab:streetview_actions}, and the execution of the returned tool call. The terminal \texttt{GUESS} action consumes one action from the budget. The \textbf{action budget} is the maximum number of actions permitted in a round. A model may terminate early by emitting \texttt{GUESS}, and is force-terminated with its current best guess once the budget is exhausted. This makes the exploration setup a guess refinement procedure.

We limit exploration to 8 steps as a principled choice reflecting realistic agentic deployment constraints. Appendix~\ref{app:actions} shows that mean Haversine distance plateaus and occasionally degrades beyond action 5, indicating that agents either commit to a hypothesis early or fail to update it meaningfully. An unconstrained budget does not help LLMs improve  (Table~\ref{tab:model_comparison_percent}) and instead amplifies confirmation bias (Appendix~\ref{app:actions}) and inflates API costs (Appendix~\ref{app:cost_analysis}) without added diagnostic value.

We initially collected a list of 100 cities across the globe based on two criteria: developed/developing status (according to the UN Council for Trade and Development\footnote{\url{https://unctadstat.unctad.org/EN/Classifications.html}}) and recognizability tiers (as outlined in Table~\ref{tab:tier_dataset_composition}). We classify cities based on annual visitor counts\footnote{\url{https://www.statista.com/chart/12742/the-worlds-most-visited-cities}}. This ensures a difficulty classification that closely resembles web-scale training data characteristics for most modern LLMs. For instance, urban agglomerates like New York City and Paris are ranked in Tier 1, while Butare, Rwanda, and Imphal, India are placed in Tier 5.

For each city, we randomly sample 100 latitude and longitude pairs within a 2-mile radius of their known coordinates as retrieved from Google's Geocoding API, resulting in a dataset of 10,000 records. Coordinates that result in no navigable Street View environments are dropped, and we finally retain 1,200 samples in our dataset with an unbalanced bifurcation as shown in Table~\ref{tab:tier_dataset_composition}. This ensures the validity and difficulty of our data, as it would resist solutions from pure pattern matching or training data regurgitation and instead force fine-grained geospatial reasoning. A sample of the collected data is shown in Figure~\ref{fig:example_data}. A total of nine tool calls are defined for the model to interact with the environment, as shown in Table~\ref{tab:streetview_actions}.


\subsection{Evaluation Metrics}

We evaluate agent performance using three complementary metrics that capture geolocation accuracy, exploration effectiveness, and reasoning patterns.

\paragraph{Geospatial Distance.}
We measure geolocation quality using the great-circle (Haversine) distance between predicted and ground-truth coordinates. Given latitude/longitude pairs $(\phi_1, \lambda_1)$ (prediction) and $(\phi_2, \lambda_2)$ (ground truth) in radians:
\begin{align}
    a &= \sin^2\left(\frac{\Delta \phi}{2}\right) + \cos(\phi_1)\cos(\phi_2)\sin^2\left(\frac{\Delta \lambda}{2}\right), \\
    c &= 2\arctan2(\sqrt{a}, \sqrt{1-a}), \\
    d &= R \cdot c,
\end{align}
where $\Delta\phi = \phi_2 - \phi_1$, $\Delta\lambda = \lambda_2 - \lambda_1$, and $R = 6371$ km is Earth's radius. We report both mean and median distances, as the distribution is highly skewed by catastrophic failures.

\paragraph{Improvement Rate.}
To quantify the exploration effect, we measure how much the agent's guess improves from its initial prediction to action $k$:
\begin{equation}
    \label{eq:improvement}
    \text{improvement}_k = \left(1 - \frac{d_k}{d_1}\right) \times 100,
\end{equation}
where $d_1$ is the distance at action 1 (initial guess) and $d_k$ is the distance at action $k$. Positive values indicate the agent moved closer to the correct location. We compute this separately for rounds with correct versus incorrect final city guesses to analyze exploration patterns.

\paragraph{TF-IDF Vocabulary Analysis.}
We identify characteristic vocabulary used by VLMs about different geographic categories using term frequency-inverse document frequency (TF-IDF). For each round, we concatenate the \texttt{observations} and \texttt{reasoning} fields from all actions. We preprocess the resulting text by: (i) lowercasing, (ii) removing non-alphabetic characters, (iii) removing English stop words, (iv) extracting unigrams and bigrams (\texttt{ngram\_range=(1,2)}), and (v) filtering terms appearing in $>95\%$ of documents (\texttt{max\_df=0.95}).

All extracted text segments from the same country are concatenated into a single document. For instance, the "Japan" document contains all observation and reasoning text from rounds where the ground-truth location was in Japan. Finally, we compute top-$N$ TF-IDF terms per country.

\subsection{Baselines}

To compare against agentic exploration, we establish \textbf{4-view and 8-view multi-view baselines} that stitch several shots of the environment into a single image for direct inference. The 4-view captures a panorama at 0°, 90°, 180°, and 270°; the 8-view adds a second panorama taken after one MOVE\_FORWARD action (examples in Appendix~\ref{app:multiview_baselines}). Further, to isolate what LLM-guided exploration contributes, we add two conditions:

\begin{itemize}
    \setlength{\itemsep}{2pt}
    \setlength{\parskip}{0pt}
    \item \textbf{Random Baseline:} A floor condition that samples a latitude/longitude pair uniformly at random for each round and submits it as the guess. It takes zero actions, produces no reasoning trace, and exists only to bound our measurements. 
    
    \item \textbf{Random Walk:} The agent explores under the same 8-action budget as our primary limited agent setting, but each action is drawn entirely at random from Table~\ref{tab:streetview_actions} rather than chosen by the model. The VLM still sees every rendered viewport and still produces the final guess. This condition helps us hold visual coverage and inference budget fixed while removing goal-directed action selection. Further, we conduct an agentic-minus-random-walk gap analysis to understand performance differences due to exploration policy.
\end{itemize}


Finally, to test the effect of the exploration budget, an \textbf{unlimited-step agent} removes the 8-action limit and allows the VLM to gather as much evidence as it wants (detailed in Appendix~\ref{app:unlimited-step}). 

\subsection{Human Reference Point}
\label{sec:human-reference}

We recruited six volunteer annotators with a mix of beginner and intermediate GeoGuessr exposure, who completed 380 rounds in total on stratified subsets of the 1{,}200-sample pool used for the model evaluations. This allowed us to control for annotation quality and per-annotator workload. Annotators played through the same Street View interface used by the agents. Further breakdown and per-annotator analysis can be found in Appendix~\ref{app:per-annotator}. We note that between-annotator variance is large (mean distance ranges from 1{,}720 to 3{,}751~km) based on previous GeoGuessr experience. Readers must treat this baseline as a reference point that supports the claim that frontier VLMs exceed \emph{non-expert} human performance on this task.

\begin{table*}[t]
\centering
\scriptsize
\vspace{0.5em}
\begin{tabular}{llcccccc}
\toprule
\textbf{Condition} & \textbf{Model} & \textbf{Mean Dist.} & \textbf{Median Dist.} & \textbf{Reas. Len.} & \textbf{Cont. Acc (\%)} & \textbf{Country Acc (\%)} & \textbf{City Acc (\%)} \\
\midrule
\multicolumn{2}{l}{\textit{Human Baseline}} & 2753.6 \scriptsize{$\pm$ 202.2} & 995.9 \scriptsize{$\pm$ 253.4} & -- & 57.6 \scriptsize{$\pm$ 2.5} & 36.6 \scriptsize{$\pm$ 2.5} & 7.1 \scriptsize{$\pm$ 1.3} \\

\midrule

& Claude Haiku 4.5 & 2951.7 \scriptsize{$\pm$ 48.9} & 1114.6 \scriptsize{$\pm$ 27.4} & 329.9 \scriptsize{$\pm$ 1.6} & 64.4 \scriptsize{$\pm$ 0.2} & 43.5 \scriptsize{$\pm$ 0.1} & 9.4 \scriptsize{$\pm$ 0.2} \\
& GPT-5 Mini & \textbf{776.7} \scriptsize{$\pm$ 10.9} & \underline{232.5} \scriptsize{$\pm$ 0.1} & 205.9 \scriptsize{$\pm$ 0.1} & \textbf{90.3} \scriptsize{$\pm$ 0.7} & \textbf{80.4} \scriptsize{$\pm$ 0.6} & \underline{30.3} \scriptsize{$\pm$ 0.4} \\
& Gemini 3.0 Flash & \underline{1446.8} \scriptsize{$\pm$ 74.8} & \textbf{118.5} \scriptsize{$\pm$ 11.7} & 232.3 \scriptsize{$\pm$ 1.9} & 70.0 \scriptsize{$\pm$ 0.5} & \underline{65.6} \scriptsize{$\pm$ 0.6} & \textbf{39.5} \scriptsize{$\pm$ 1.0} \\
\multirow{-4}{*}{\rotatebox[origin=c]{90}{\textbf{Static 4-View}}} & Gemma 3 27B & 2669.2 \scriptsize{$\pm$ 18.0} & 754.7 \scriptsize{$\pm$ 0.5} & \textbf{198.0} \scriptsize{$\pm$ 0.2} & 49.5 \scriptsize{$\pm$ 0.3} & 37.1 \scriptsize{$\pm$ 0.2} & 12.7 \scriptsize{$\pm$ 0.2} \\
& Llama 4 Scout & 1586.1 \scriptsize{$\pm$ 61.2} & 458.0 \scriptsize{$\pm$ 3.7} & \underline{200.4} \scriptsize{$\pm$ 0.3} & 72.7 \scriptsize{$\pm$ 0.3} & 59.9 \scriptsize{$\pm$ 0.7} & 20.8 \scriptsize{$\pm$ 0.3} \\
& Geo-R1 7B & 2350.1 \scriptsize{$\pm$ 59.0} & 521.0 \scriptsize{$\pm$ 17.8} & 226.4 \scriptsize{$\pm$ 1.6} & \underline{74.9} \scriptsize{$\pm$ 0.3} & 62.8 \scriptsize{$\pm$ 0.6} & 17.5 \scriptsize{$\pm$ 0.2} \\
& \textcolor{gray}{Random Baseline} & \textcolor{gray}{9958.2 \scriptsize{$\pm$ 73.8}} & \textcolor{gray}{9951.2 \scriptsize{$\pm$ 164.9}} & \textcolor{gray}{--} & \textcolor{gray}{0.0} & \textcolor{gray}{0.0} & \textcolor{gray}{0.0} \\

\midrule

& Claude Haiku 4.5 & 2603.2 \scriptsize{$\pm$ 39.6} & 944.7 \scriptsize{$\pm$ 36.3} & 338.6 \scriptsize{$\pm$ 2.1} & 68.7 \scriptsize{$\pm$ 0.4} & 47.9 \scriptsize{$\pm$ 0.6} & 9.9 \scriptsize{$\pm$ 0.6} \\
& GPT-5 Mini & \textbf{740.1} \scriptsize{$\pm$ 44.0} & \underline{205.7} \scriptsize{$\pm$ 6.5} & \textbf{200.5} \scriptsize{$\pm$ 0.3} & \textbf{91.5} \scriptsize{$\pm$ 1.1} & \textbf{82.9} \scriptsize{$\pm$ 1.2} & \underline{32.3} \scriptsize{$\pm$ 0.6} \\
& Gemini 3.0 Flash & \underline{1247.3} \scriptsize{$\pm$ 34.1} & \textbf{84.1} \scriptsize{$\pm$ 5.2} & 231.9 \scriptsize{$\pm$ 1.3} & 71.9 \scriptsize{$\pm$ 0.7} & \underline{67.3} \scriptsize{$\pm$ 0.8} & \textbf{41.5} \scriptsize{$\pm$ 0.5} \\
\multirow{-4}{*}{\rotatebox[origin=c]{90}{\textbf{Static 8-View}}} & Gemma 3 27B & 3092.0 \scriptsize{$\pm$ 12.7} & 888.7 \scriptsize{$\pm$ 6.7} & 205.1 \scriptsize{$\pm$ 0.3} & 44.3 \scriptsize{$\pm$ 0.4} & 33.4 \scriptsize{$\pm$ 0.3} & 12.0 \scriptsize{$\pm$ 0.3} \\
& Llama 4 Scout & 1592.6 \scriptsize{$\pm$ 36.7} & 453.5 \scriptsize{$\pm$ 8.9} & \underline{203.2} \scriptsize{$\pm$ 0.4} & 75.0 \scriptsize{$\pm$ 0.4} & 61.5 \scriptsize{$\pm$ 0.7} & 20.3 \scriptsize{$\pm$ 0.8} \\
& Geo-R1 7B & 2154.4 \scriptsize{$\pm$ 31.0} & 439.8 \scriptsize{$\pm$ 3.0} & 226.6 \scriptsize{$\pm$ 0.2} & \underline{77.4} \scriptsize{$\pm$ 0.4} & 67.0 \scriptsize{$\pm$ 0.6} & 18.7 \scriptsize{$\pm$ 0.4} \\
& \textcolor{gray}{Random Baseline} & \textcolor{gray}{9992.1 \scriptsize{$\pm$ 193.1}} & \textcolor{gray}{9951.7 \scriptsize{$\pm$ 268.0}} & \textcolor{gray}{--} & \textcolor{gray}{0.0} & \textcolor{gray}{0.0} & \textcolor{gray}{0.0} \\

\midrule

& Claude Haiku 4.5 & 3260.4 \scriptsize{$\pm$ 30.0} & 1031.9 \scriptsize{$\pm$ 50.9} & \underline{710.2} \scriptsize{$\pm$ 2.3} & 49.7 \scriptsize{$\pm$ 0.3} & 43.2 \scriptsize{$\pm$ 0.5} & 11.7 \scriptsize{$\pm$ 0.7} \\
& GPT-5 Mini & \underline{1581.3} \scriptsize{$\pm$ 50.3} & \underline{283.8} \scriptsize{$\pm$ 17.1} & \underline{239.5} \scriptsize{$\pm$ 1.7} & \textbf{73.0} \scriptsize{$\pm$ 0.7} & \textbf{71.2} \scriptsize{$\pm$ 0.6} & \underline{30.0} \scriptsize{$\pm$ 1.3} \\
& Gemini 3.0 Flash & \textbf{1565.1} \scriptsize{$\pm$ 484.3} & \textbf{184.5} \scriptsize{$\pm$ 199.9} & 337.8 \scriptsize{$\pm$ 43.4} & \underline{62.2} \scriptsize{$\pm$ 9.6} & \underline{64.7} \scriptsize{$\pm$ 12.3} & \textbf{35.1} \scriptsize{$\pm$ 12.4} \\
\multirow{-4}{*}{\rotatebox[origin=c]{90}{\textbf{Random Walk}}} & Gemma 3 27B & 2205.3 \scriptsize{$\pm$ 37.7} & 434.9 \scriptsize{$\pm$ 14.8} & 395.7 \scriptsize{$\pm$ 0.3} & 49.5 \scriptsize{$\pm$ 0.3} & 48.1 \scriptsize{$\pm$ 0.4} & 20.2 \scriptsize{$\pm$ 1.3} \\
& Llama 4 Scout & 2782.5 \scriptsize{$\pm$ 55.6} & 701.5 \scriptsize{$\pm$ 21.9} & 340.3 \scriptsize{$\pm$ 1.4} & 52.9 \scriptsize{$\pm$ 0.8} & 44.9 \scriptsize{$\pm$ 0.5} & 16.2 \scriptsize{$\pm$ 0.3} \\
& Geo-R1 7B & 2404.5 \scriptsize{$\pm$ 38.9} & 521.0 \scriptsize{$\pm$ 8.2} & \textbf{229.9} \scriptsize{$\pm$ 0.6} & 56.9 \scriptsize{$\pm$ 0.0} & 53.8 \scriptsize{$\pm$ 0.7} & 9.2 \scriptsize{$\pm$ 0.2} \\
& \textcolor{gray}{Random Baseline} & \textcolor{gray}{7956.1 \scriptsize{$\pm$ 3652.4}} & \textcolor{gray}{6741.7 \scriptsize{$\pm$ 5838.9}} & \textcolor{gray}{--} & \textcolor{gray}{0.0} & \textcolor{gray}{0.0} & \textcolor{gray}{0.0} \\

\midrule

& Claude Haiku 4.5 & 2181.4 \scriptsize{$\pm$ 86.9} & 688.3 \scriptsize{$\pm$ 32.9} & 424.2 \scriptsize{$\pm$ 2.0} & 68.4 \scriptsize{$\pm$ 2.6} & 52.4 \scriptsize{$\pm$ 2.7} & 14.3 \scriptsize{$\pm$ 1.3} \\
& GPT-5 Mini & \textbf{900.0} \scriptsize{$\pm$ 73.5} & \underline{182.3} \scriptsize{$\pm$ 11.6} & \textbf{182.8} \scriptsize{$\pm$ 0.7} & \textbf{88.6} \scriptsize{$\pm$ 0.2} & \textbf{81.2} \scriptsize{$\pm$ 0.7} & \underline{34.5} \scriptsize{$\pm$ 1.4} \\
& Gemini 3.0 Flash & \underline{1099.4} \scriptsize{$\pm$ 99.8} & \textbf{5.0} \scriptsize{$\pm$ 0.2} & 283.5 \scriptsize{$\pm$ 27.3} & \underline{73.8} \scriptsize{$\pm$ 1.2} & \underline{69.8} \scriptsize{$\pm$ 1.4} & \textbf{50.1} \scriptsize{$\pm$ 0.5} \\
\multirow{-4}{*}{\rotatebox[origin=c]{90}{\shortstack{\textbf{Limited}\\\textbf{Agentic Navigation}}}} & Gemma 3 27B & 1689.3 \scriptsize{$\pm$ 147.5} & 401.5 \scriptsize{$\pm$ 48.2} & 268.5 \scriptsize{$\pm$ 100.4} & 63.9 \scriptsize{$\pm$ 1.9} & 54.0 \scriptsize{$\pm$ 3.1} & 19.3 \scriptsize{$\pm$ 1.7} \\
& Llama 4 Scout & 1779.5 \scriptsize{$\pm$ 35.5} & 579.7 \scriptsize{$\pm$ 12.7} & \underline{200.4} \scriptsize{$\pm$ 0.2} & 70.1 \scriptsize{$\pm$ 1.2} & 58.4 \scriptsize{$\pm$ 1.3} & 14.7 \scriptsize{$\pm$ 1.0} \\
& Geo-R1 7B & 2549.0 \scriptsize{$\pm$ 188.2} & 575.8 \scriptsize{$\pm$ 94.8} & 224.1 \scriptsize{$\pm$ 4.3} & 61.8 \scriptsize{$\pm$ 1.9} & 54.6 \scriptsize{$\pm$ 1.6} & 17.9 \scriptsize{$\pm$ 1.6} \\
& \textcolor{gray}{Random Baseline} & \textcolor{gray}{10089.2 \scriptsize{$\pm$ 94.2}} & \textcolor{gray}{10210.3 \scriptsize{$\pm$ 184.1}} & \textcolor{gray}{--} & \textcolor{gray}{0.0} & \textcolor{gray}{0.0} & \textcolor{gray}{0.0} \\

\midrule

& Claude Haiku 4.5 & 3162.0 \scriptsize{$\pm$ 250.5} & 972.8 \scriptsize{$\pm$ 146.0} & 978.9 \scriptsize{$\pm$ 18.1} & 67.4 \scriptsize{$\pm$ 3.6} & 52.1 \scriptsize{$\pm$ 2.0} & 12.8 \scriptsize{$\pm$ 1.0} \\
& GPT-5 Mini & \textbf{1136.1} \scriptsize{$\pm$ 78.0} & \underline{182.0} \scriptsize{$\pm$ 31.9} & 564.9 \scriptsize{$\pm$ 8.2} & \textbf{86.9} \scriptsize{$\pm$ 1.0} & \textbf{80.2} \scriptsize{$\pm$ 0.3} & \underline{35.8} \scriptsize{$\pm$ 1.7} \\
& Gemini 3.0 Flash & \underline{1229.2} \scriptsize{$\pm$ 435.6} & \textbf{149.8} \scriptsize{$\pm$ 204.8} & \underline{517.7} \scriptsize{$\pm$ 58.8} & \underline{84.5} \scriptsize{$\pm$ 7.3} & \underline{76.5} \scriptsize{$\pm$ 10.4} & \textbf{39.5} \scriptsize{$\pm$ 15.3} \\
\multirow{-4}{*}{\rotatebox[origin=c]{90}{\shortstack{\textbf{Unlimited}\\\textbf{Agentic Navigation}}}} & Gemma 3 27B & 1610.8 \scriptsize{$\pm$ 222.9} & 441.0 \scriptsize{$\pm$ 72.5} & 595.8 \scriptsize{$\pm$ 3.8} & 73.4 \scriptsize{$\pm$ 4.0} & 59.3 \scriptsize{$\pm$ 3.2} & 20.1 \scriptsize{$\pm$ 2.3} \\
& Llama 4 Scout & 3136.5 \scriptsize{$\pm$ 231.5} & 680.7 \scriptsize{$\pm$ 32.8} & 523.6 \scriptsize{$\pm$ 8.3} & 69.3 \scriptsize{$\pm$ 3.4} & 59.3 \scriptsize{$\pm$ 4.5} & 20.7 \scriptsize{$\pm$ 1.7} \\
& Geo-R1 7B & 2846.6 \scriptsize{$\pm$ 81.1} & 559.0 \scriptsize{$\pm$ 35.1} & \textbf{352.7} \scriptsize{$\pm$ 2.8} & 66.7 \scriptsize{$\pm$ 1.0} & 63.1 \scriptsize{$\pm$ 1.5} & 19.1 \scriptsize{$\pm$ 1.2} \\
& \textcolor{gray}{Random Baseline} & \textcolor{gray}{10089.2 \scriptsize{$\pm$ 94.2}} & \textcolor{gray}{10210.3 \scriptsize{$\pm$ 184.1}} & \textcolor{gray}{--} & \textcolor{gray}{0.0} & \textcolor{gray}{0.0} & \textcolor{gray}{0.0} \\

\bottomrule

\end{tabular}
\caption{Performance comparison across conditions and models. \textbf{Bold} = best, \underline{underline} = second-best per condition. Accuracies reported in \%. Lower distance and reasoning length are better; higher accuracy is better.}
\label{tab:model_comparison_percent}
\end{table*}

\subsection{Models Used}

We test six leading VLMs on our task. Among open options, we choose Gemma 3 27B and Llama 4 Scout 109B with Claude Haiku 4.5, Gemini 3.0 Flash, and GPT-5 Mini forming the closed subset. We test all models through their official APIs, other than Llama 4, which we access through Groq. Geo-R1 7B is a specialized spatial reasoning model and is hosted locally on an Nvidia RTX 5070 Ti. We set the temperature to 0.3 when available.

\begin{figure*}[t]
    \centering
    \includegraphics[width=0.95\textwidth]{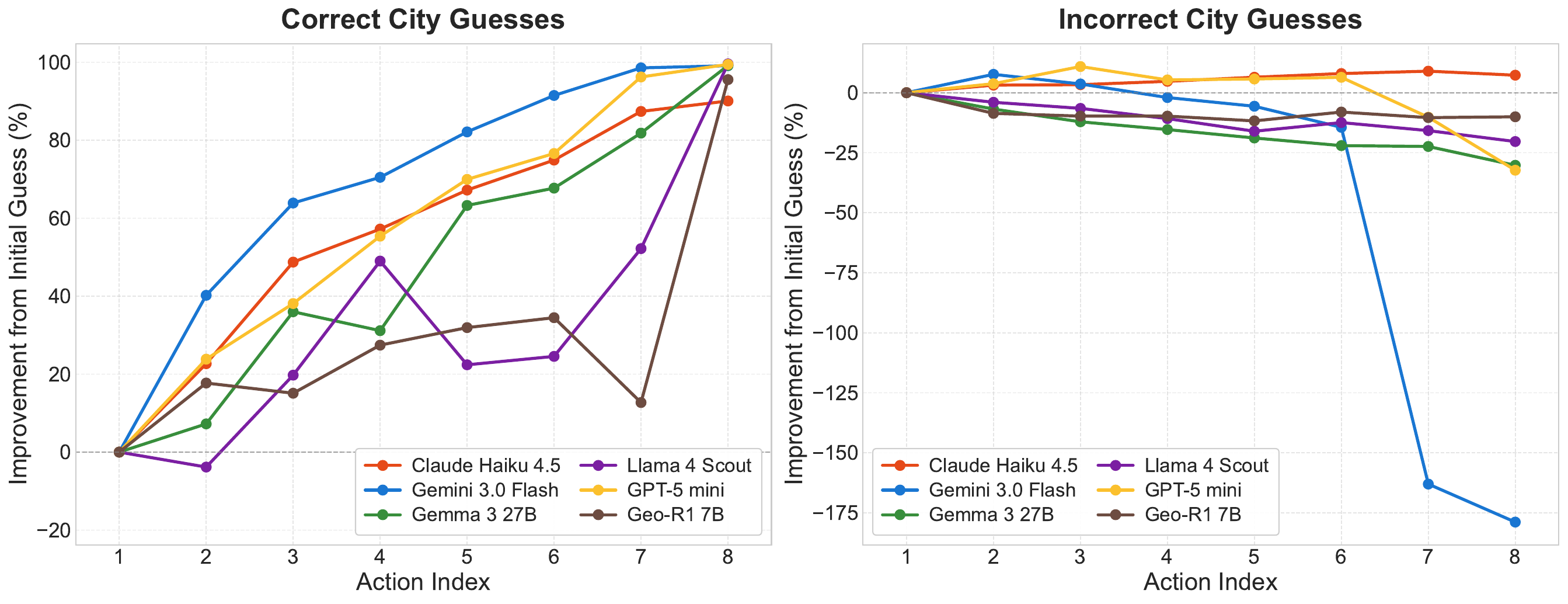}
    \caption{Aggregated performance improvement per action (out of 8) per model, stratified by final correct and incorrect guesses. We notice model performance generally improves with each extra action, but calibrates when the outcome is incorrect. Gemini 3.0 Flash is an outlier with a collapse on incorrect city guesses beyond 5 actions.}
    \label{fig:action_comparison}
\end{figure*}

\section{Results}
\label{sec:results}

Table~\ref{tab:model_comparison_percent} summarizes our main results, where Gemini 3.0 Flash takes the lead over the remaining models (median Haversine distance of 5 km and 50.1\% city prediction accuracy) with Claude Haiku 4.5 performing the worst (688.3 km median Haversine distance). Interestingly, GPT-5 Mini outperforms Gemini on mean Haversine distance (900 km), continent accuracy (88.6\%), and country accuracy (81.2\%). Notably, all models show significantly improved performance over the human reference point (995.9 km median Haversine distance, 7.1\% city accuracy). Claude Haiku 4.5 falls below the human reference on mean distance under the 4-view ($2951.7$~km), random-walk ($3260.4$~km), and unlimited ($3162.0$~km) settings, as do Llama 4 Scout and Geo-R1 7B under unlimited exploration. Unlimited exploration fails to provide any meaningful gains despite its significantly higher cost (Appendix~\ref{app:cost_analysis}), further motivating the proposition of the limited GeoAgent. The specialized Geo-R1 7B is competitive on the static settings (mean distance 2350.1 / 2154.4 km on 4-view / 8-view; continent accuracy up to 77.4\%), but is the only model whose agentic performance \emph{degrades} relative to its static baselines (2549.0 km mean distance, 61.8\% continent accuracy) as it emits invalid tool calls in 87.1\% turns. 99.8\% if its navigation is not effective (Appendix~\ref{app:geor1-forensics}).

We also notice higher standard deviations on agentic navigation against the 4-view and 8-view baselines, but the numbers remain contextualized to the random baseline. The substantial gap between country-level accuracy (52.4–81.2\%) and city-level accuracy (14.3–50.1\%) across all models suggests that VLMs can identify broad geographic regions from macro-level cues (vegetation, climate indicators, driving side) but struggle with the "last-mile" localization based on municipal infrastructure, local business signage, or regional architectural variants. Gemma 3 27B demonstrates strong performance, beating both Claude Haiku 4.5 and Llama 4 Scout in city prediction accuracies and Haversine distance despite its smaller size. 

Interestingly, across model averages, we observe that GPT-5 Mini is the strongest model on distance yet produces the least verbose traces (182.8 words), while Claude Haiku 4.5 is the weakest and the most verbose (424.2 words). However, at the level of individual rounds pooled across all runs, this relationship reverses and is statistically weakly positive (Appendix~\ref{app:stat_validity}). Finally, this remains just an observation as there's no causal reading in which writing style makes a model more accurate.

In comparison to the 4-view and 8-view baselines, we notice significantly improved geolocalization performance across both Haversine distance ($1446.79\pm74.82$ km on 4-view to 1099.38 km on agentic for Gemini 3.0 Flash) and city-level accuracy, which sees the largest relative gains: Gemini 3.0 Flash improves from 39.5\% (4-view) to 50.1\% (agentic), a 26.8\% relative increase, while GPT-5 Mini rises from 30.3\% to 34.5\%. The random-walk condition isolates the contribution of LLM-guided exploration, as every general-purpose VLM improves substantially over its random-walk counterpart on mean distance (GPT-5 Mini $1581.3 \to 900.0$ km, Gemma 3 27B $2205.3 \to 1689.3$ km, Claude Haiku 4.5 $3260.4 \to 2181.4$ km). Interestingly, Gemini 3.0 Flash shows large standard deviations in this setting, which is fully mitigated with agentic navigation. This suggests that step-wise exploration allows models to iteratively refine coarse regional guesses. Notably, not all models benefit equally from exploration: Llama 4 Scout's mean distance actually worsens from 1586.08 km (4-view) to 1779.45 km (agentic). 

\subsection{Improving VLM geoguessing performance through embodied navigation}

\begin{figure*}
    \centering
    \includegraphics[width=0.93\textwidth]{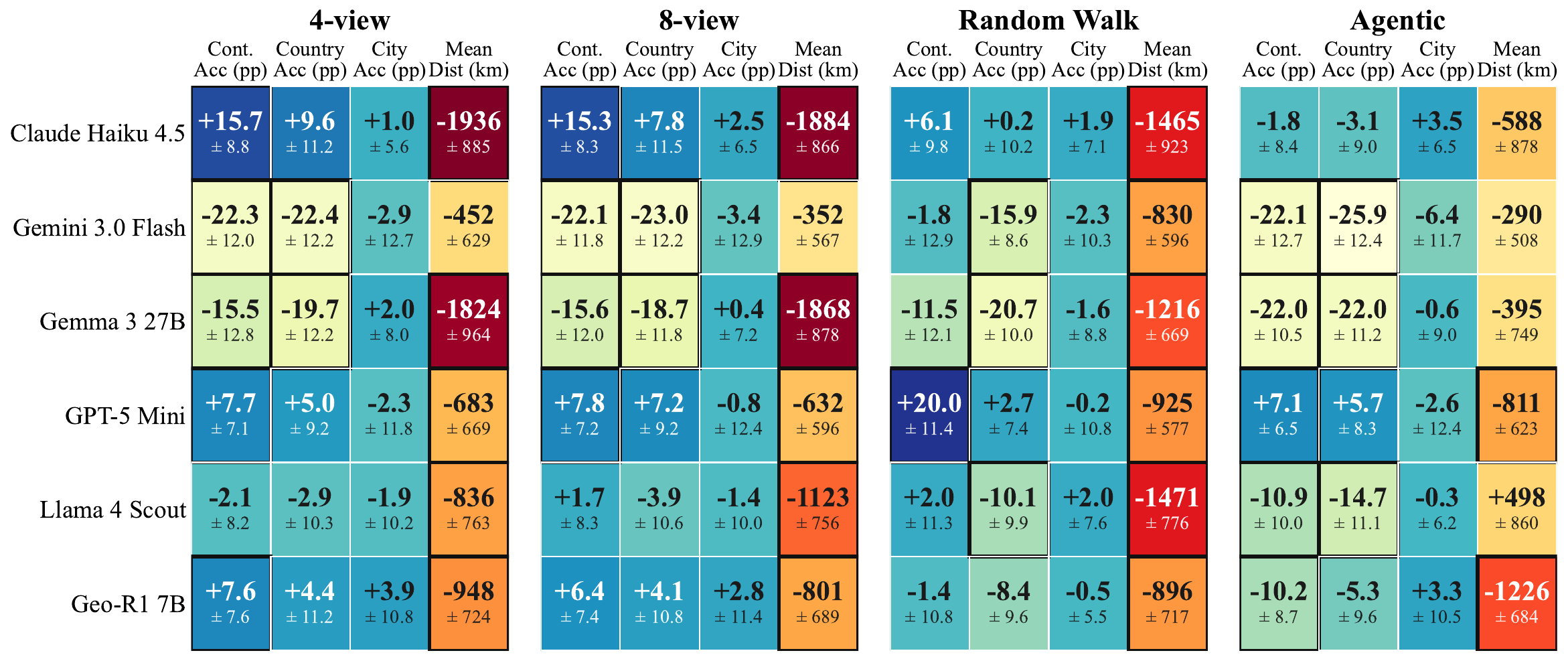}
    \caption{Developed–developing delta across all models and navigation modes. Bordered cells exclude zero in their confidence intervals. Each cell shows the difference (developed --developing); negative values indicate stronger performance on developed regions. All models exhibit a developed-region bias with 50–103\% lower mean distance, and most secure higher accuracies for developed regions. Gemma 3 27B is an exception, with slightly higher city accuracy in developing regions across all three conditions. Gemini 3.0 Flash shows the largest country-level deltas (>20\%) in all modes, suggesting its strong performance is disproportionately driven by developed-region familiarity.} 
    \label{fig:improvement_comparison}
\end{figure*}


\begin{table*}[t]
\centering
\scriptsize
\setlength{\tabcolsep}{6pt}
\renewcommand{\arraystretch}{1.15}
\begin{tabular}{lccccccc}
\toprule
\multirow{2}{*}{\textbf{Model}} &
\multicolumn{4}{c}{\textbf{Exploration behavior}} &
\multicolumn{3}{c}{\textbf{Mean Haversine distance (km)}} \\
\cmidrule(lr){2-5}\cmidrule(lr){6-8}
 & \textbf{Actions} & \textbf{Moves} & \textbf{Rotations} & \textbf{Invalid (\%)}
 & \textbf{Agentic} & \textbf{Random Walk} & \textbf{$\Delta$} \\
\midrule
GPT-5 Mini        & 4.94{\tiny$\pm$0.07} & 2.42{\tiny$\pm$0.12} & 1.51{\tiny$\pm$0.06} & 0.00
                  & \textbf{900}{\tiny$\pm$74}  & 1581{\tiny$\pm$50}  & $+681$ \\
Gemini 3.0 Flash  & 4.96{\tiny$\pm$0.25} & 3.57{\tiny$\pm$0.25} & 0.39{\tiny$\pm$0.02} & 0.00
                  & 1099{\tiny$\pm$100} & 1565{\tiny$\pm$484} & $+466$ \\
Claude Haiku 4.5  & 5.61{\tiny$\pm$0.49} & 2.79{\tiny$\pm$0.36} & 1.59{\tiny$\pm$0.14} & 0.04
                  & 2181{\tiny$\pm$87}  & 3260{\tiny$\pm$30}  & $\mathbf{+1079}$ \\
Gemma 3 27B       & 5.96{\tiny$\pm$0.25} & 4.20{\tiny$\pm$0.34} & 0.76{\tiny$\pm$0.49} & 0.00
                  & 1689{\tiny$\pm$148} & 2205{\tiny$\pm$37}  & $+516$ \\
Llama 4 Scout     & 7.55{\tiny$\pm$0.04} & 0.41{\tiny$\pm$0.07} & 6.14{\tiny$\pm$0.07} & 0.00
                  & 1779{\tiny$\pm$36}  & 2783{\tiny$\pm$55}  & $+1004$ \\
Geo-R1 7B         & 7.85{\tiny$\pm$0.22} & 0.00{\tiny$\pm$0.00} & 0.00{\tiny$\pm$0.00} & \textbf{87.3}
                  & 2549{\tiny$\pm$188} & 2405{\tiny$\pm$38}  & $\mathbf{-144}$ \\
\bottomrule
\end{tabular}
\caption{Exploration behavior and the causal contribution of goal-directed navigation under the 8-action budget. \textbf{Actions} is the mean number of turns used before the agent commits; \textbf{Moves} and \textbf{Rotations} are the mean counts of each per round; \textbf{Invalid} is the share of emitted actions the environment could not parse. $\Delta$ is random-walk minus agentic mean distance, so positive values mean model-chosen navigation beats random movement. Every model with functioning tool use gains 466--1079~km; Geo-R1 7B, which emits an unparseable action in 87.3\% of turns and never navigates, is the sole model that does worse than random walk (Appendix~\ref{app:geor1-forensics}).}
\label{tab:exploration_behavior}
\end{table*}

To study whether additional navigation improves geoguessing capabilities, we quantify agent behavior over each action in Figure~\ref{fig:action_comparison}. For guesses that are finally correct, we notice that most models show significant performance improvement with more navigation. For incorrect guesses, however, most models show limited sensitivity to additional exploration. Gemini 3.0 Flash, interestingly, shows an exponential drop in accuracy with added navigation steps beyond action 5 (Figure~\ref{fig:action_comparison}, right). Claude Haiku 4.5, in contrast, shows the most calibration, as added navigation does not significantly improve or worsen its guesses. While exploration improves performance when initial hypotheses are approximately correct (RQ1), VLMs demonstrate poor self-correction capabilities. Additional actions rarely recover from incorrect priors, suggesting confirmation bias in sequential reasoning. Importantly, removing the action cap budget does not restore recovery (Appendix~\ref{app:unlimited-step}).

\begin{figure*}
    \centering
    \includegraphics[width=0.95\textwidth]{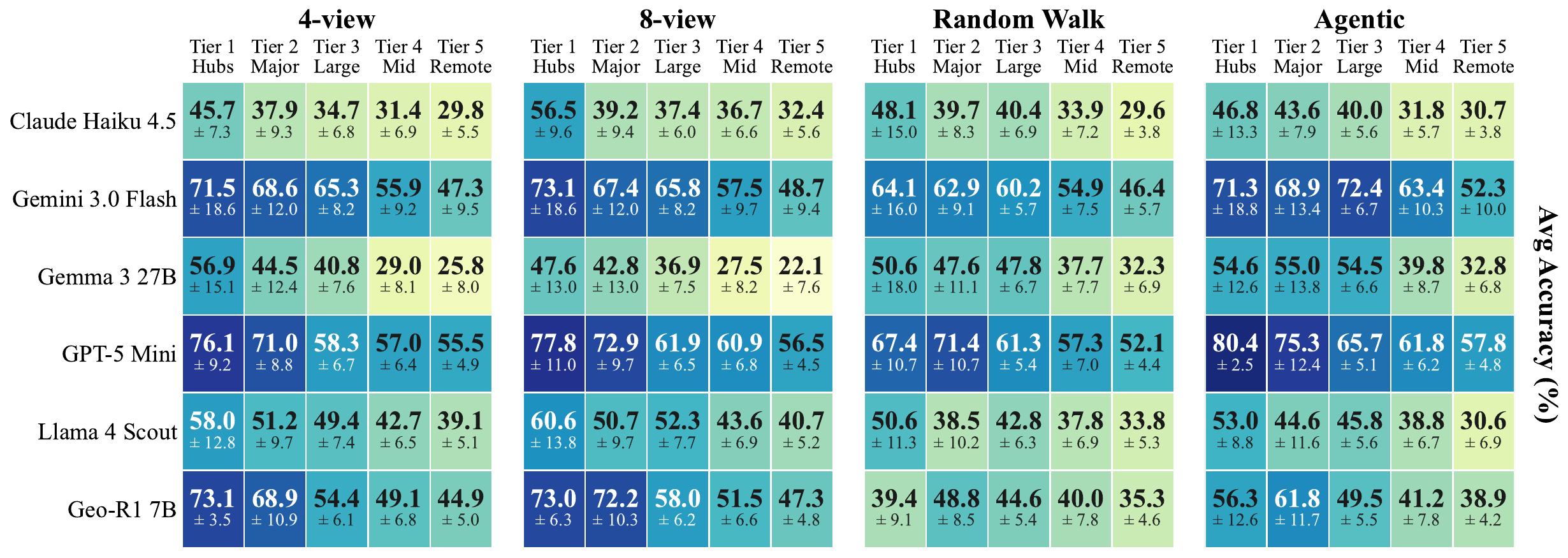}
    \caption{Agent behavior and localization accuracy across geographic tiers. Models struggle significantly with less identifiable locations, with city guessing accuracies dropping below 26\% on Tier 5. Interestingly, Llama 4 Scout and GPT-5 Mini maintain their country guessing accuracies across tiers while Gemma3 and Haiku 4.5 show collapse down the tier list. Other than Llama 4 Scout, mean Haversine distance generally increases by tier.}
\label{fig:tier_behavior_accuracy}
\end{figure*}

We further investigate what distinguishes recoverable from non-recoverable initial errors by analyzing the relationship between first-action distance and final accuracy. For rounds where models ultimately succeed, the median initial distance is 847 km (IQR: 312–1,892 km), compared to 2,341 km (IQR: 1,156–4,872 km) for failed rounds. A logistic regression predicting final city accuracy from initial distance yields a negative coefficient ($\beta=-0.43, p<0.001$), confirming that exploration refines already-reasonable hypotheses instead of re-discovery. Our statistical tests (Appendix~\ref{app:stat_validity}) show that these results are significant.


\subsection{Agent Behavior Analysis}


Overall, VLMs are significantly better at guessing developed region locations than developing alternatives (Figure~\ref{fig:improvement_comparison}). Notably, we find that models explore developed more than developing regions (5.95 vs. 5.65 actions), and reasoning length is largely architecture-dependent. When models do explore developing-region locations fully, they exhibit higher rotation-to-move ratios (2.1:1 vs. 1.4:1 in developed regions). While this difference is not directly noticed in task accuracy (on average, developed regions lead by 2.07\%), Haversine distance deviates by 50-103\%, representing substantial bias across the tested models. The largest deltas appear at the country level under agentic navigation: Gemini 3.0 Flash shows a -25.9\% country accuracy gap and GPT-5 Mini a -5.7\% gap, while Claude Haiku 4.5 and Llama 4 Scout show -3.0\% and -14.7\% respectively. Gemma 3 27B is a notable exception, with slightly higher city accuracy in developing regions across all three conditions (Figure~\ref{fig:improvement_comparison}). Similar results are also noted in a city population tier-based analysis (Figure~\ref{fig:tier_behavior_accuracy}). 

To better understand the agentic behavior of the tested VLMs, we dissect performance by development tiers and recognizability tiers. Table~\ref{tab:exploration_behavior} shows how the number of exploration actions, defined as a sum of moves and rotations made by the model, differs, and whether agentic navigation helps it. Addressing RQ3, we note that GPT-5 Mini is quite efficient, completing its task in 4.20-4.95 steps out of 8. Llama 4 Scout shows higher utilization at 7.48-7.57 steps, but most of this remains concentrated to rotations across the spawn point instead of moving forward or backward.  Our statistical results in Appendix~\ref{app:stat_validity} show that taking more screenshots helps up to a point, then plateaus or slightly hurts city-level precision. Geo-R1 7B, despite being a specialized model, struggles at exploration, which leads to slight performance degradation with agentic navigation (further forensics in Appendix~\ref{app:geor1-forensics}). 


\subsection{Reasoning Vocabulary Analysis}


Finally, we conduct TF-IDF sampling to identify which visual cues and terminology are the most frequently encountered in geospatial reasoning. Finding more qualitative evidence for RQ2, we note that models focus on visible identification cues across tested nations, such as "gingko" for Japan, "tram tracks" and "shutters" for Switzerland, and the "Chao Phraya" river for Thailand. Most models reason with nationality-based terminology across the regions tested; however, a few noticeable patterns emerge. In Claude Haiku 4.5, we notice that "Japanese" is the leading term for Taiwan, and Gemma 3 27B leads with "German" for Switzerland. Table~\ref{tab:agentic_vocab_overlap} shows that models share 36.6-52.7\% of their frequent vocabulary across the tested settings, meaning this is an LLM-specific behavior and not an artifact of the GeoAgent framework. Qualitatively, we notice false positives cluster around visually similar regions: Southeast Asian countries are frequently confused (Thai locations misclassified as Vietnamese 23\% of the time, and vice versa 18\%). Notably, models often invoke the correct visual descriptors ("tropical vegetation," "weathered infrastructure," and "motorcycle traffic" for Southeast Asia), but map these to incorrect conclusions, suggesting fine-grained geographic grounding remains missing. Countries with the highest misclassification rates (Mozambique 67\%, Thailand 54\%) have no country-specific vocabulary in their reasoning chains at all, suggesting models fall back on generic regional descriptors when they lack distinctive cues. Targeted data augmentation with country-specific visual-textual pairs could help underperforming regions.

\begin{table}
\centering
\scriptsize
\setlength{\tabcolsep}{8pt}
\renewcommand{\arraystretch}{1.2}
\begin{tabular}{lcc}
\toprule
\textbf{Model} & \textbf{Agentic $\leftrightarrow$ 4-View} & \textbf{Agentic $\leftrightarrow$ 8-View} \\
\midrule
Claude Haiku 4.5        & 41.7\% & 39.0\% \\
Gemini 3.0 Flash         & 52.7\% & 51.9\% \\
Gemma 3 27B          & 37.2\% & 36.6\% \\
Llama 4 Scout  & 39.7\% & 43.2\% \\
GPT-5 mini     & 50.0\% & 48.5\% \\
\bottomrule
\end{tabular}
\caption{Agentic vocabulary overlap (top 50 TF-IDF words) with static multiview settings. Reported values are the mean across all geographic categories.}
\label{tab:agentic_vocab_overlap}
\end{table}

\section{Discussion}

First, active exploration can improve geolocation performance, but the improvements are highly conditional (\textbf{RQ1}). When agents begin with a reasonably good initial hypothesis (i.e., the initial guess is ``in the right neighborhood''), additional navigation tends to refine that hypothesis and substantially reduce Haversine distance (see Figure~\ref{fig:improvement_comparison} and Appendix~\ref{app:actions}). However, for many trials where the initial guess is far off, extra exploration often fails to recover and sometimes reinforces incorrect priors. This suggests that VLMs are often better at confirmation than recovery, and established mistakes in context largely influence outcomes.

Second, our analyses uncover a marked and systematic bias favoring developed-region locations (\textbf{RQ2}). Although continent and country accuracies remain relatively high, mean Haversine distances and failure modes are substantially worse for developing-region samples (Table~\ref{tab:exploration_behavior} and Figure~\ref{fig:tier_behavior_accuracy}), and the bias also surfaces in the regional vocabulary models rely on (Appendix~\ref{app:tf-idf-tables}). This is a model behavior artifact, as developing regions have newer and marginally higher quality data per Google's quality score. They are also indistinguishable from developed regions in sharpness (Appendix~\ref{app:data-quality}). Third, exploration length and reasoning verbosity do not directly correlate to higher success (Appendix~\ref{app:stat_validity}). 

Fourth, model architecture and pretraining regimes strongly affect how efficiently agents use the limited action budget (\textbf{RQ3}). Gemini 3.0 Flash and GPT-5 Mini achieved the best overall accuracies and the lowest median distances while using substantially fewer actions on average (Tables~\ref{tab:model_comparison_percent} and~\ref{tab:exploration_behavior}). In contrast, models such as Llama 4 Scout used many rotations but few forward moves. Moreover, Geo-R1 7B is limited by its tool calling capabilities, and its performance does not improve with agentic navigation. This suggests that good static-image geolocation performance does not directly translate into effective embodied exploration across frontier VLMs, and performance gains with agentic navigation are conditional on basic tool calling capabilities.

\section{Conclusion}

We introduce GeoAgent, a geolocation guessing task for VLMs that require embodied agentic web exploration. Our analyses reveal several behavioral caveats in current-generation models that are overlooked by benchmarks focusing on static image-based recognition: lack of self-correction when initial guesses are incorrect, poor efficiency tuning in models such as Claude Haiku 4.5 and Llama 4 Scout, and a significant developed/developing bias. Through our experiments, we argue for improved pre-training for embodied geospatial navigation and location predictions with alternate evaluation angles not studied in previous literature.

\section*{Limitations}

There are several limitations to acknowledge. First, although we evaluate an unlimited-action variant with reflection prompts (Appendix~\ref{app:unlimited-step}), our headline numbers use a fixed 8-action budget. Mean Haversine distance plateaus beyond action 5 (Appendix~\ref{app:actions}) and the unlimited setup yields no consistent gains across the three models we re-ran, but a per-model adaptive budget could surface qualitatively different exploration regimes we leave to future work. Second, our action space (nine actions, with continuous heading control via \texttt{ROTATE:<degrees>}) mirrors Street View's underlying capture topology but does not capture every real-world navigation strategy (e.g., zoom, multi-step planning, or cross-modal lookup). Third, our human baseline draws on six annotators with beginner-to-experienced GeoGuessr exposure (Appendix~\ref{app:per-annotator}); while it clearly establishes that frontier VLMs exceed non-expert human performance, an expert-player comparison would strengthen the human-mimicry framing further. Fourth, while our sample of 1{,}200 locations spans 100 designed cities across six continents (Appendix~\ref{app:datasheet}), it is not intended as a comprehensive geographic benchmark. The value of GeoAgent lies in its agentic, embodied evaluation paradigm rather than absolute geographic coverage.

Finally, GeoAgent's dependence on Google Street View and the Google Geocoding API raises a reproducibility concern. This is particularly data in the current corpus drifts, and Google can alter it without notice. Moreover, both APIs are commercial, rate-limited, and subject to terms that permit deprecation. However, we must note the limitations of open-source alternatives such as OpenStreetMap\footnote{https://www.openstreetmap.org}, which does not allow for a principled study with detailed navigable environments. To promote reproducibility, our released code and data include the resolved \texttt{pano\_id}, capture date, heading, and pitch alongside the target coordinate, besides full per-action logs.

\section*{Acknowledgments}

TIH IIT Tirupati (IITTNiF/TPD/202425/P16) partially supported this research work. We also thank Nvidia India for providing Blackwell-based RTX 50-series GPUs to run the experiments. Finally, we thank the six volunteer annotators who contributed the human reference point.

\section*{Ethical Considerations}

Geolocalization systems like GeoAgent raise legitimate concerns around privacy and potential misuse, like surveillance applications and unauthorized location inference from personal media. GeoAgent does not process user-generated content and operates exclusively on publicly available Google Street View imagery, which blurs all privately identifiable information, such as license plates and faces. However, we acknowledge that advances in agentic geolocalization could lower the barrier for misuse if applied to unconstrained image sources. Additionally, our dataset inherits the well-documented geographic coverage bias of Street View, which disproportionately represents developed regions with higher-quality imagery, potentially amplifying performance disparities across nations. We encourage future work to incorporate explicit misuse safeguards, such as restricted API access and geofencing, alongside more geographically balanced data collection efforts.



\bibliography{custom}

\clearpage
\appendix

\noindent{\LARGE \textbf{Appendices}}
\vspace{0.5cm}

This supplementary material presents additional details on the following aspects:  
\begin{itemize}
    \setlength{\itemsep}{0pt}
    \setlength{\parsep}{0pt}
    \setlength{\topsep}{0pt}
    \setlength{\leftmargin}{1em}
    \item \textbf{Appendix~\ref{app:prompts}:} Prompts
    \item \textbf{Appendix~\ref{app:actions}:} Distance Convergence Across Actions
    \item \textbf{Appendix~\ref{app:stat_validity}:} Action-Outcome Correlation Analysis
    \item \textbf{Appendix~\ref{app:multiview_baselines}:} Multiview Baseline Implementation
    \item \textbf{Appendix~\ref{app:unlimited-step}:} Unlimited Action Step Exploration
    \item \textbf{Appendix~\ref{app:improvement_analysis}:} $d_1$ to $d_k$ Improvement for Small $d_1$
    \item \textbf{Appendix~\ref{app:cost_analysis}:} Cost and Efficiency Analysis
    \item \textbf{Appendix~\ref{app:data-quality}:} Data Quality Analysis
    \item \textbf{Appendix~\ref{app:geor1-forensics}:} Why Geo-R1 Degrades Under Navigation
    \item \textbf{Appendix~\ref{app:subregion}:} Subregion Breakdown and Retention
    \item \textbf{Appendix~\ref{app:per-annotator}:} Per-Annotator Breakdown of Human Baselines
    \item \textbf{Appendix~\ref{app:tf-idf-tables}:} Full TF-IDF Tables
    \item \textbf{Appendix~\ref{app:datasheet}:} Detailed Datasheet
\end{itemize}

\section{Prompts}
\label{app:prompts}

Figure~\ref{fig:geoguessr_prompt} presents the prompt template used across all evaluated VLMs. The prompt structure consists of four components: (1) task framing, establishing the GeoGuessr context, (2) dynamic state information including round number, remaining actions, and current heading, (3) contextual memory of up to five prior observations from the current round, and (4) the available action space.

We enforce structured JSON output to ensure consistent parsing across model architectures. Each response requires five fields: observations (salient visual cues), reasoning (geographic deduction), confidence level (four-point scale), selected action, and current location guess with city and country. The Google Geocoding API is then used to deduce the latitude and longitude for the guessed location to avoid any LLM-induced hallucination in coordinate calculations. Models are encouraged to keep the Observations and reasoning fields brief, as verbose outputs correlate negatively with task performance (see Appendix~\ref{app:stat_validity}).

\begin{figure}[h]
\centering
\fbox{%
\begin{minipage}{0.97\columnwidth}
\footnotesize
\textbf{GeoGuessr Agent Prompt}

\vspace{0.3em}
You are playing \textit{GeoGuessr}. Guess the location from the image.

\vspace{0.3em}
\textbf{State:}
Round \texttt{\{round\_num\}} \;
|\; Actions left \texttt{\{max\_actions - count\}} \;
|\; Heading \texttt{\{heading\}$^\circ$}

\vspace{0.3em}
\textbf{Context:}
Up to the last five observations from the current round, if available.

\vspace{0.3em}
\textbf{Available actions:}
\texttt{ROTATE\_LEFT},
\texttt{ROTATE\_RIGHT},
\texttt{ROTATE\_BEHIND},
\texttt{MOVE\_FORWARD},
\texttt{RETURN\_START},
\texttt{GUESS}

\vspace{0.3em}
\textbf{Response format (JSON only):}

{\ttfamily\footnotesize
\begin{quote}
\{
 "observations": "Brief key clues",

 "reasoning": "Short deduction",

 "confidence": "low|medium|high|very\_high",

 "action": "ROTATE\_*|MOVE\_*|GUESS",

 "guess": \{
   "location": "Place, City, Country",
   "country": "Country or null",
   "city": "City or null"
 \}
\}
\end{quote}
}
\end{minipage}}
\vspace{-0.5em}
\caption{Prompt used by the GeoGuessr agent for location inference and action selection.}
\label{fig:geoguessr_prompt}
\vspace{-0.8em}
\end{figure}

\section{Distance Convergence Across Actions}
\label{app:actions}

\begin{figure*}
    \centering
    \includegraphics[width=\textwidth]{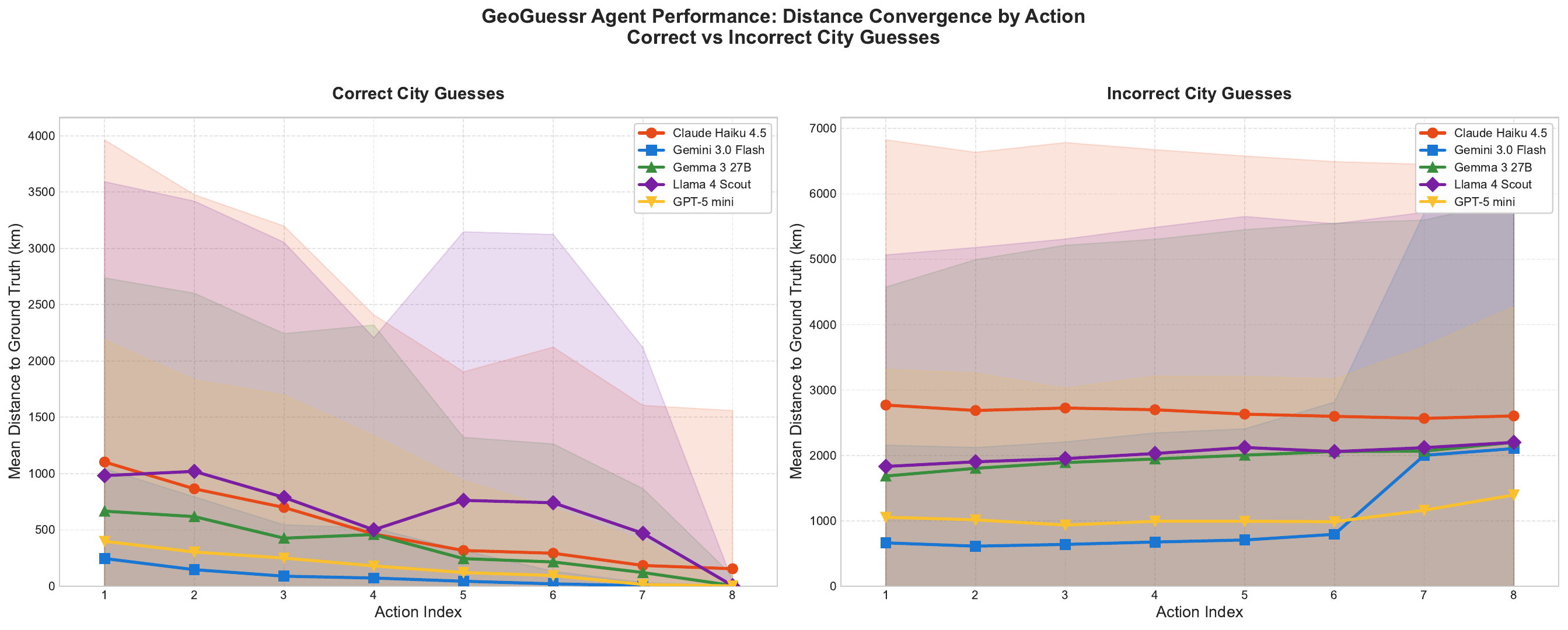}
    \caption{Aggregated mean distance improvement over actions per model.}
    \label{fig:dist_comparison}
\end{figure*}

Figure~\ref{fig:dist_comparison} presents the mean Haversine distance to ground truth across sequential actions, stratified by outcome correctness. For rounds culminating in correct city guesses (left panel), all models demonstrate monotonic convergence toward the target location, with GPT-5 Mini achieving sub-500 km distances by action 3. Notably, Llama 4 Scout and Claude Haiku 4.5 exhibit high initial uncertainty (>1500 km), but only the former converges effectively given sufficient exploration budget.

The right panel reveals that when final guesses are incorrect, models show minimal distance improvement regardless of exploration effort. Claude Haiku 4.5 and Gemma 3 27B maintain relatively stable (but incorrect) estimates, while GPT-5 Mini diverges after action 5. This asymmetry provides evidence for confirmation bias in sequential geospatial reasoning, where models struggle to abandon initial misconceptions even when presented with contradictory visual evidence.

\section{Action-Outcome Correlation Analysis }
\label{app:stat_validity}

Table~\ref{tab:behavior_correlation} reports Pearson and Spearman correlations between agent behavioral features and task outcomes across all model runs. We observe statistically significant positive correlations between reasoning length and all accuracy measures (continent: $\rho = 0.140$, country: $\rho = 0.137$, city: $\rho = 0.090$), alongside a negative correlation with distance ($\rho = -0.142$), suggesting that longer reasoning traces are associated with more accurate geolocalization rather than mere verbosity. Reasoning length shows the strongest individual association with points ($\rho = 0.241$), reinforcing that deliberate multi-step analysis contributes to task performance.

Total actions and screenshots taken also correlate positively with continent and country accuracy ($\rho = 0.124$ and $\rho = 0.118$ for total actions), though the relationship weakens considerably at the city level (Pearson $r = -0.009$, Spearman $\rho = 0.055$), indicating that additional exploration aids broad localization but offers diminishing returns for fine-grained disambiguation. Move actions show a consistent positive association across all outcomes, with the strongest link to points ($\rho = 0.195$) and distance ($\rho = -0.108$), suggesting that forward movement is a more productive exploration strategy than rotation. This aligns with our earlier finding that Llama 4 Scout's rotation-heavy policy (6.14 rotations vs.\ 0.41 moves per round) underperforms relative to Gemini 3.0 Flash's movement-dominant strategy (3.57 moves vs.\ 0.39 rotations). All correlations reject the null hypothesis at $\alpha = 0.05$, supporting the finding that increased exploration effort correlates with improved geospatial reasoning, though effect sizes remain modest ($|\rho| < 0.25$), indicating that action quality continues to matter more than action quantity.

\section{Multiview Baseline Implementation}
\label{app:multiview_baselines}

Figure 
represents a sample from the dataset curated for evaluation of the VLM for the Multiview-Baseline. 
The multiview baseline is a non-agentic pipeline in which the VLM is provided with a single, static composite image for each location, and performs exactly one inference, without interactive navigation or iterative reasoning.

\textbf{Composite Capture}. We use a fully automated script, producing deterministic composite images for all 1200 locations. 
For each location, the street view panorama is first positioned to the target coordinates. Four screenshots are captured at offsets of  0°, 90°, 180°, and 270° from the initial heading. To capture the 8-view composite, the system additionally performs a \texttt{MOVE\_FORWARD} action by selecting the navigation link with the closest heading alignment, after which four more screenshots are taken with identical offsets. 
The eight views are arranged into a labeled 4×2 grid composite (2560×776 px), with original spawn-point views occupying the top row and post-move views the bottom. A 4-view spawn-only composite is produced similarly.
 Both are saved as JPEG and accompanied by a manifest recording per-location metadata including coordinates, development status, and recognizability tier.

 \begin{figure}[!t]
\centering
\fbox{%
\begin{minipage}{0.97\columnwidth}
\footnotesize
\textbf{Multiview Baseline Prompt}
\vspace{0.3em}

You are playing \textit{GeoGuessr}. Guess the location from the image.
\vspace{0.3em}

\textbf{State:}
Round \texttt{\{N\}}
\vspace{0.3em}

\textbf{Response format (JSON only):}
{\ttfamily\footnotesize
\begin{quote}
\{
 "observations": "Brief key clues,

 "reasoning": "Short deduction",

 "confidence": "low|medium|high|very\_high",

 "guess": \{
   "location": "Specific place, City, Country",
   "country": "Country or null",
   "city": "City or null"
 \}
\}
\end{quote}
}

\end{minipage}}
\vspace{-0.5em}
\caption{Prompt used by the multiview baseline models for single-pass location inference. Unlike the agentic prompt (Figure~\ref{fig:geoguessr_prompt}), no action selection, remaining action count, or heading state is provided, reflecting the non-interactive nature of the baseline.}
\label{fig:multiview_prompt}
\vspace{-0.8em}
\end{figure}

 \textbf{Offline Inference}. The pre-captured composite is passed to each model alongside the prompt in Figure ~\ref{fig:multiview_prompt}. The prompt omits state and action fields. All other prompt structure, output format requirements, temperature settings, and geocoding procedures are kept identical to the agentic condition. This ensures that performance differences between the agentic and multiview conditions are attributable to embodied navigation rather than prompt or scoring asymmetries.

\begin{figure*}
\centering
    \includegraphics[width=\textwidth]{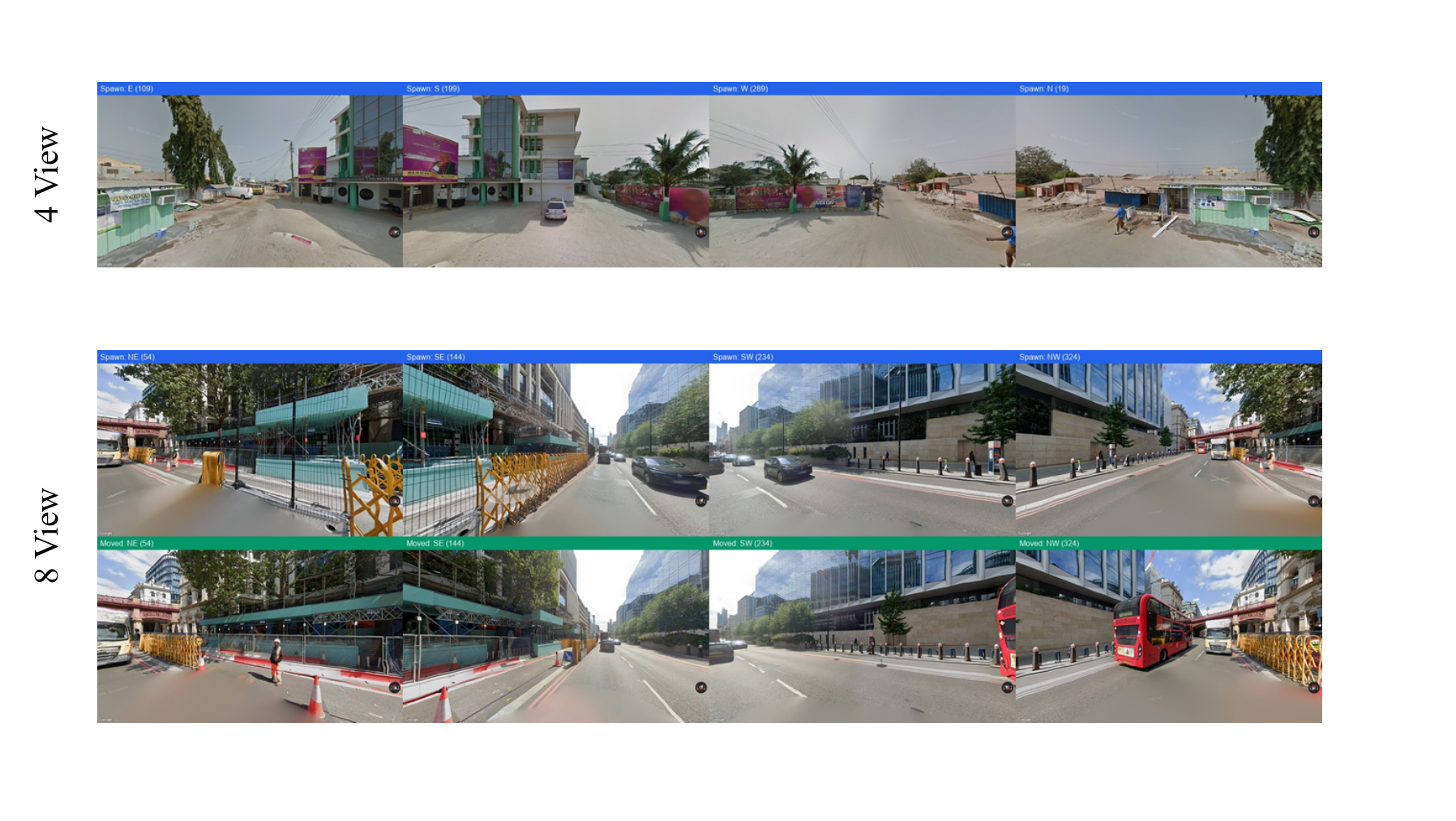}
    \caption{Examples of 4-view and 8-view composite images used in the multiview baseline. The top strip shows four spawn-point screenshots at cardinal headings (4-view). The bottom two rows show the full 8-view composite, with spawn-point views (blue labels) in the first row and post-\texttt{MOVE\_FORWARD} views (green labels) in the second row, each captured at identical heading offsets.}
\label{fig:multiview_composites}
\end{figure*}

\begin{table*}[ht]
\centering
\small
\setlength{\tabcolsep}{6pt}
\renewcommand{\arraystretch}{1.25}
\begin{tabular}{llccccc}
\toprule
\textbf{Feature} &
\textbf{Outcome} &
\textbf{Pearson $r$} &
\textbf{Pearson $p$} &
\textbf{Spearman $\rho$} &
\textbf{Rejects Null Hypothesis ($H_0$)?} \\
\midrule
\multirow{5}{*}{\shortstack{Reasoning\\Length}}
& Continent Correct & $0.129$  & $0.0000$ & $0.140$  & YES \\
& Country Correct   & $0.119$  & $0.0000$ & $0.137$  & YES \\
& City Correct      & $0.070$  & $0.0000$ & $0.090$  & YES \\
& Distance Km       & $-0.130$ & $0.0000$ & $-0.142$ & YES \\
& Points            & $0.181$  & $0.0000$ & $0.241$  & YES \\
\midrule
\multirow{5}{*}{\shortstack{Screenshots\\Taken}}
& Continent Correct & $0.097$  & $0.0000$ & $0.128$  & YES \\
& Country Correct   & $0.081$  & $0.0000$ & $0.121$  & YES \\
& City Correct      & $-0.009$ & $0.0213$ & $0.057$  & YES \\
& Distance Km       & $-0.085$ & $0.0000$ & $-0.102$ & YES \\
& Points            & $0.121$  & $0.0000$ & $0.216$  & YES \\
\midrule
\multirow{5}{*}{\shortstack{Total\\Actions}}
& Continent Correct & $0.096$  & $0.0000$ & $0.124$  & YES \\
& Country Correct   & $0.081$  & $0.0000$ & $0.118$  & YES \\
& City Correct      & $-0.009$ & $0.0167$ & $0.055$  & YES \\
& Distance Km       & $-0.085$ & $0.0000$ & $-0.108$ & YES \\
& Points            & $0.120$  & $0.0000$ & $0.211$  & YES \\
\midrule
\multirow{5}{*}{\shortstack{Move\\Actions}}
& Continent Correct & $0.099$  & $0.0000$ & $0.122$  & YES \\
& Country Correct   & $0.086$  & $0.0000$ & $0.114$  & YES \\
& City Correct      & $0.014$  & $0.0004$ & $0.047$  & YES \\
& Distance Km       & $-0.099$ & $0.0000$ & $-0.108$ & YES \\
& Points            & $0.140$  & $0.0000$ & $0.195$  & YES \\
\bottomrule
\end{tabular}
\caption{Correlation analysis between agent behavior features and task outcomes. Pearson and Spearman correlation coefficients are reported along with hypothesis test results.}
\label{tab:behavior_correlation}
\end{table*}

\begin{table*}[h!]
\centering
\scriptsize
\setlength{\tabcolsep}{3pt}
\renewcommand{\arraystretch}{1.2}
\begin{tabular}{llcccccccc}
\toprule
\textbf{Model} & 
\multirow{2}{*}{\textbf{\shortstack{Dev.\\Level}}} &
\multirow{2}{*}{\textbf{\shortstack{\#\\Actions}}} &
\multirow{2}{*}{\textbf{\shortstack{\#\\Moves}}} &
\multirow{2}{*}{\textbf{\shortstack{\#\\Rotations}}} &
\multirow{2}{*}{\textbf{\shortstack{Mean\\Reas. Len.}}} &
\multicolumn{3}{c}{\textbf{Accuracy (\%)}} & 
\multirow{2}{*}{\textbf{\shortstack{Mean Dist.\\(km)}}} \\
\cmidrule(lr){7-9}
 &  &  &  &  &  & \textbf{Continent} & \textbf{Country} & \textbf{City} &  \\
\midrule
\multirow{2}{*}{\shortstack{Random\\Baseline}}
& Developed  & - & - & - & - & $0.00 \pm 0.0$ & $0.00 \pm 0.0$ & $0.00 \pm 0.0$ & $9982.18 \pm 306.7$ \\
& Developing & - & - & - & - & $0.00 \pm 0.0$ & $0.00 \pm 0.0$ & $0.00 \pm 0.0$ & $10007.06 \pm 255.4$ \\
\midrule
\multirow{2}{*}{\shortstack{Claude\\Haiku 4.5}}
& Developed  & $6.29 \pm 0.1$ & $3.15 \pm 0.1$ & $2.04 \pm 0.1$ & $434.7 \pm 7.3$ & $68.23 \pm 6.9$ & $53.22 \pm 4.4$ & $17.93 \pm 1.7$ & $1782.64 \pm 263.2$ \\
& Developing & $5.40 \pm 0.0$ & $2.68 \pm 0.1$ & $1.64 \pm 0.0$ & $415.4 \pm 8.9$ & $61.53 \pm 1.4$ & $43.92 \pm 4.9$ & $14.08 \pm 1.4$ & $3712.72 \pm 410.9$ \\
\midrule
\multirow{2}{*}{\shortstack{Gemma 3\\27B}}
& Developed  & $6.92 \pm 0.1$ & $2.70 \pm 0.2$ & $3.22 \pm 0.2$ & $220.0 \pm 2.0$ & $45.94 \pm 2.7$ & $36.35 \pm 1.1$ & $22.81 \pm 0.5$ & $1307.74 \pm 315.4$ \\
& Developing & $6.76 \pm 0.0$ & $3.89 \pm 0.1$ & $1.87 \pm 0.1$ & $221.7 \pm 0.9$ & $71.18 \pm 1.8$ & $56.56 \pm 1.5$ & $17.92 \pm 3.1$ & $2536.77 \pm 114.8$ \\
\midrule
\multirow{2}{*}{\shortstack{Llama 4\\Scout}}
& Developed  & $7.57 \pm 0.1$ & $0.20 \pm 0.0$ & $6.37 \pm 0.1$ & $197.0 \pm 3.1$ & $64.53 \pm 2.0$ & $49.07 \pm 2.5$ & $14.75 \pm 2.1$ & $1622.49 \pm 107.6$ \\
& Developing & $7.48 \pm 0.1$ & $0.44 \pm 0.0$ & $6.04 \pm 0.1$ & $205.4 \pm 1.7$ & $73.89 \pm 3.3$ & $61.06 \pm 1.7$ & $14.69 \pm 2.5$ & $2472.46 \pm 363.1$ \\
\midrule
\multirow{2}{*}{\shortstack{GPT-5\\mini}}
& Developed  & $4.95 \pm 0.5$ & $2.46 \pm 0.4$ & $1.49 \pm 0.1$ & $183.4 \pm 1.4$ & $91.69 \pm 2.0$ & $83.72 \pm 2.1$ & $34.42 \pm 2.4$ & $645.95 \pm 163.7$ \\
& Developing & $4.20 \pm 0.3$ & $2.14 \pm 0.3$ & $1.07 \pm 0.1$ & $182.0 \pm 1.9$ & $84.62 \pm 1.0$ & $76.44 \pm 1.8$ & $34.93 \pm 1.1$ & $1316.75 \pm 291.6$ \\
\bottomrule
\end{tabular}
\caption{Agent behavior and accuracy breakdown across developed and developing regions. Models generally make similar exploration actions across both region types, but the final outcomes vary widely. Most models secure higher continent, country, and city guessing accuracies, while Gemma 3 27B reverses this trend. Importantly, all tested VLMs show an acute developed region bias in mean Haversine distance, resulting in 50-103\% better numbers than the equivalent developing region result (lower is better).}
\label{tab:behavior_accuracy}
\end{table*}

\section{Unlimited Action Step Exploration}
\label{app:unlimited-step}

To test whether the 8-action cap masks recoverable failure modes -- i.e., whether the confirmation-bias signal in \S\ref{app:actions} is an artefact of an exhausted action budget -- we re-run three models (Gemma 3 27B, Llama 4 Scout, GPT-5 Mini) under an unlimited-action variant with explicit self-reflection. Models reason and act linearly for the first five actions; at the sixth action, we inject a reflection prompt asking the model to reconsider its exploration so far and either submit a guess or continue. If the model elects to continue, the reflection loop repeats until the model commits. Context windows are unbounded, so prior observations are never truncated.

Table~\ref{tab:geoagent_performance} reports the comparison. Performance differences are model-dependent but generally favour the limited setup. GPT-5 Mini's mean distance worsens (\,$899.99 \to 1218.41$ km) despite using more than twice as many actions per round (\,$4.94 \to 10.13$), and Llama 4 Scout shows the most pathological behavior: its action count expands threefold (\,$7.55 \to 22.99$) while its mean distance worsens by 76\% (\,$1779.45 \to 3136.53$ km). Only Gemma 3 27B improves on distance ($1689.33 \to 1577.65$ km) while keeping a nearly identical action count ($5.96 \to 5.36$), suggesting the gain is driven by the reflection prompt itself rather than by additional exploration. Across all three models, mean reasoning length roughly doubles or triples under the unlimited setup, yet continent and city accuracies move only by a few points. Together with the random-walk results in \S\ref{tab:model_comparison_percent}, these findings reinforce two claims: (i) the confirmation-bias signal in our main analysis is not an artefact of the action cap, and (ii) longer reasoning traces do not translate into more accurate localisation. We also note an order-of-magnitude increase in inference cost under the unlimited setting (Appendix~\ref{app:cost_analysis}), which motivates our pragmatic 8-action design.

\begin{table*}
\centering
\scriptsize
\setlength{\tabcolsep}{5pt}
\renewcommand{\arraystretch}{1.2}
\begin{tabular}{lcccccc}
\toprule
\textbf{Metric} &
\textbf{Gemma-3-27B} &
\textbf{Gemma-3-27B} &
\textbf{Llama-4 Scout} &
\textbf{Llama-4 Scout} &
\textbf{GPT-5-mini} &
\textbf{GPT-5-mini} \\
&
\textbf{(Unlimited Agent)} &
\textbf{(Limited Agent)} &
\textbf{(Unlimited Agent)} &
\textbf{(Limited Agent)} &
\textbf{(Unlimited Agent)} &
\textbf{(Limited Agent)} \\
\midrule

Avg Distance (km) &
1577.65$\pm$323.61 &
1689.33$\pm$147.50 &
3136.53$\pm$228.14 &
1779.45$\pm$35.50 &
1218.41$\pm$311.27 &
899.99$\pm$73.51 \\

Avg Actions / Round &
5.36$\pm$0.08 &
5.96$\pm$0.25 &
22.99$\pm$1.68 &
7.55$\pm$0.04 &
10.13$\pm$0.47 &
4.94$\pm$0.07 \\

Avg Reasoning Length &
586.40$\pm$12.12 &
268.52$\pm$100.38 &
523.63$\pm$8.27 &
200.35$\pm$0.20 &
566.30$\pm$12.73 &
182.77$\pm$0.72 \\

Continent Accuracy &
79.0$\pm$7.0 &
63.9$\pm$2.0 &
69.0$\pm$3.0 &
70.1$\pm$1.0 &
88.0$\pm$5.0 &
88.6$\pm$0.0 \\

Country Accuracy &
61.0$\pm$8.0 &
54.0$\pm$3.0 &
59.0$\pm$5.0 &
58.4$\pm$1.0 &
79.0$\pm$7.0 &
81.2$\pm$1.0 \\

City Accuracy &
18.0$\pm$1.0 &
19.3$\pm$1.7 &
21.0$\pm$2.0 &
14.7$\pm$1.0 &
37.0$\pm$1.0 &
34.5$\pm$1.0 \\

\bottomrule
\end{tabular}

\caption{GeoAgent performance of selected models in unlimited vs.\ limited action step exploration.}
\label{tab:geoagent_performance}
\end{table*}

\section{\texorpdfstring{$d_1$ to $d_k$}{d1 to dk} Improvement Analysis}
\label{app:improvement_analysis}

The improvement-rate metric in Equation~\ref{eq:improvement} is undefined at $d_1 = 0$ and, more importantly, can be dominated by very small $d_1$ values because the normalization is $1/d_1$. A model that begins essentially on top of the ground truth can produce arbitrarily large negative ``improvements'' for any subsequent action that wanders even a few hundred meters. To check whether this destabilizes our main analysis, we stress-test the metric on the bottom 1\% of $d_1$ values, pooled across the two best-performing models (GPT-5 Mini and Gemini 3.0 Flash). This slice contains $n=54$ rounds with median $d_1 = 0.28$~km.

Two observations stand out. First, the limited-action budget protects the metric: only 1 of 54 bottom-1\% rounds reaches all eight actions, and that single round terminates at $d_8 = 1{,}819$~km -- i.e., models with very-good initial guesses overwhelmingly commit early. Second, the more behaviorally meaningful diagnostic is whether the model \emph{drifts} from a good start, which we quantify as $\max_k(d_k) > 2 d_1$ across the round. This drift criterion fires in 20.4\% of bottom-1\% rounds versus 15.0\% of the full dataset -- a modest excess, not a catastrophic blow-up.

\begin{figure*}
    \centering
    \includegraphics[width=\textwidth]{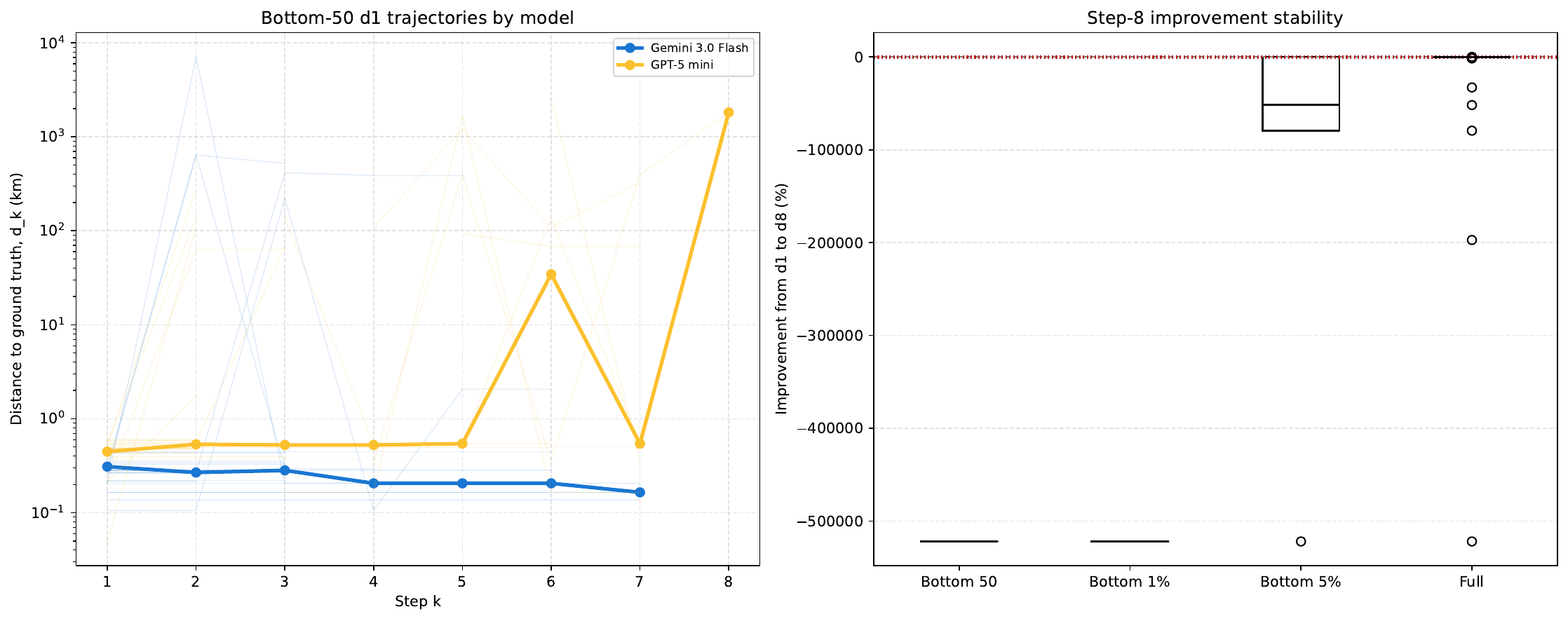}
    \caption{Model improvement behaviour under a good start (bottom-1\% of $d_1$, pooled GPT-5 Mini and Gemini 3.0 Flash; $n=54$, median $d_1 = 0.28$ km). Drift fraction ($\max_k(d_k) > 2 d_1$) is 20.4\% in this slice versus 15.0\% over the full dataset.}
    \label{fig:good_start_drift}
\end{figure*}

Because the central tendency (median improvement) is unaffected but the tails are dominated by $1/d_1$ amplification, we report \emph{median} improvement throughout the main paper and \emph{exclude rounds with $d_1 < 1$~km from any mean-based aggregate} reported in Figure~\ref{fig:action_comparison}. Figure~\ref{fig:good_start_drift} visualizes per-step distance distributions and drift fractions across all three of these stress-test models, confirming the modest-drift-but-no-blow-up pattern. We release per-step distributions, drift fractions, and per-model breakdowns in our code release.

\section{Cost and Efficiency Analysis}
\label{app:cost_analysis}

Tables~\ref{tab:model_cost_summary} and \ref{tab:agentic_cost_analysis} quantify the dollar cost of running each model across the full evaluation set (three runs over 1{,}200 samples). Two findings stand out. First, cost and task performance are sharply \emph{anti-correlated} at the top of the distribution. While Claude Haiku 4.5 is the most expensive model in the study (\$187.60 total, \$0.0174 per round) yet is also the worst-performing closed model on every accuracy metric in Table~\ref{tab:model_comparison_percent}, while Gemma 3 27B finishes the study for \$2.33 (\$0.00021 per round), an 83$\times$ cost reduction over Claude, but beats it on city-level accuracy under agentic navigation. GPT-5 Mini sits on the efficient frontier: it spends roughly one-fifth as much as Claude (\$39.43) while delivering the strongest mean Haversine distance and country accuracy. The gap is driven by Claude's verbosity (424 reasoning tokens/round, the longest in the study) compounding over an average of 5.6 actions per round.

Second, lifting the 8-action cap does not buy proportional accuracy. Table~\ref{tab:agentic_cost_analysis} shows the per-round cost of GPT-5 Mini increasing 30$\times$ (\$0.000608 $\to$ \$0.018) and Llama 4 Scout's 56$\times$ (\$0.000186 $\to$ \$0.010) under the unlimited-action variant, while continent and city accuracies move by only a few points (Appendix~\ref{app:unlimited-step}) and mean distance \emph{worsens} for both models. Gemma 3 27B is the only model where unlimited setup is cheaper in aggregate (\$0.129 vs \$0.163), because its reflection-prompted runs terminate slightly earlier. Together, these numbers justify the 8-action design choice: it preserves the benchmark's diagnostic signal at roughly an order-of-magnitude lower API spend than an unconstrained agent loop, and it does so without disadvantaging any of the evaluated model families.

\begin{table}
\centering
\scriptsize
\setlength{\tabcolsep}{5pt}
\renewcommand{\arraystretch}{1.2}
\begin{tabular}{lccc}
\toprule
\textbf{Model} & \textbf{Total Tokens} & \textbf{Total Cost (\$)} & \textbf{Cost / Round (\$)} \\
\midrule
Claude Haiku 4.5      & 42,416,725 & 187.60 & 0.0174  \\
GPT-5-mini            & 54,747,219 & 39.43  & 0.00355 \\
Llama 4 Scout  & 89,050,944 & 10.90  & 0.00098 \\
Gemini 3 Flash        & 44,421,928 & 8.50   & 0.00079 \\
Gemma 3 27B           & 23,296,767 & 2.33   & 0.00021 \\
\bottomrule
\end{tabular}

\caption{Summary of total tokens consumed, total cost in USD, and the cost per round for each model.}
\label{tab:model_cost_summary}
\end{table}

\begin{table*}[t]
\centering
\small
\setlength{\tabcolsep}{6pt}
\renewcommand{\arraystretch}{1.2}
\begin{tabular}{lcccccc}
\toprule
\textbf{Model} & \textbf{Actions/round} & \textbf{Limited} & \textbf{Limited} & \textbf{Actions/round} & \textbf{Unlimited} & \textbf{Unlimited} \\
&
\textbf{for Limited} & \textbf{\$/round} & \textbf{Total \$} & \textbf{for Unlimited} & \textbf{\$/round} & \textbf{Total \$} \\
\midrule
Gemma 3 27B & 5.96 $\pm$ 0.25 & \$0.00004525 & \$0.1629 & 5.36 $\pm$ 0.08 & \$0.000431 & \$0.1292 \\
GPT-5 Mini & 4.94 $\pm$ 0.07 & \$0.00060825 & \$2.1897 & 10.13 $\pm$ 0.47 & \$0.018215 & \$5.4644 \\
Llama 4 Scout & 7.55 $\pm$ 0.04 & \$0.00018642 & \$0.6713 & 22.99 $\pm$ 1.68 & \$0.010409 & \$3.1226 \\
\bottomrule
\end{tabular}

\caption{Cost analysis of limited and unlimited agentic exploration.}
\label{tab:agentic_cost_analysis}
\end{table*}

\section{Data Quality Analysis}
\label{app:data-quality}

\begin{figure*}[htbp]
    \centering
    \includegraphics[width=\textwidth]{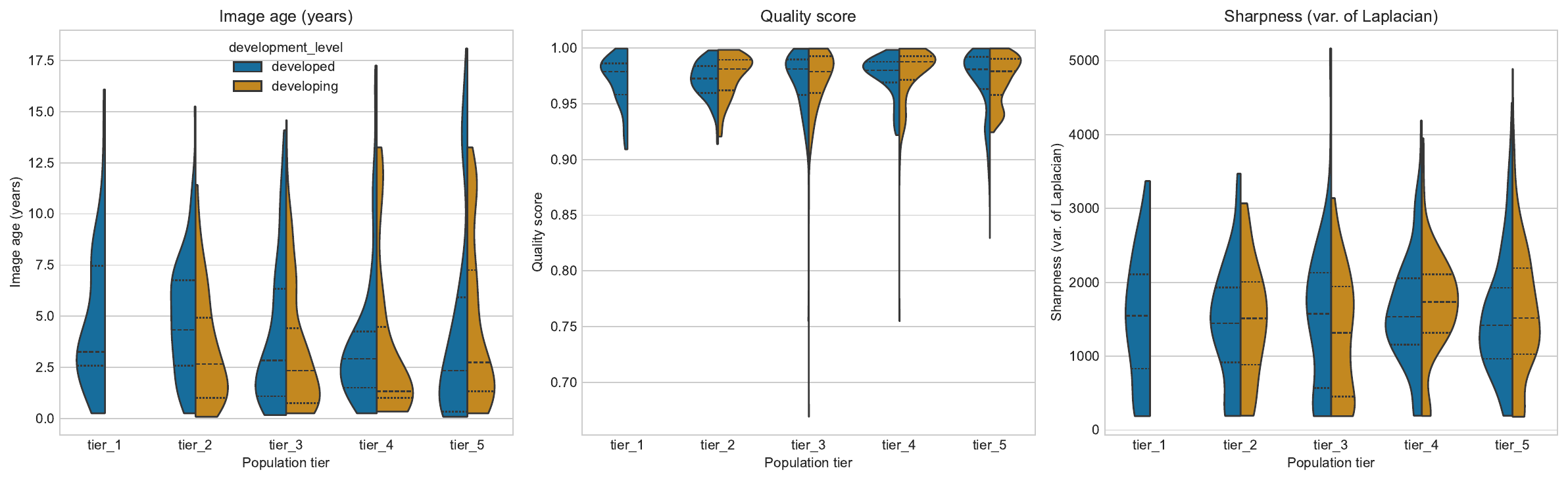}
    \caption{Per-tier and per-development-level distributions of image age, quality score, and sharpness ($n = 1{,}200$). No monotonic degradation is observed from Tier 1 to Tier 5.}
    \label{fig:coverage_confound}
\end{figure*}

A natural concern with the developed/developing performance gap reported in the main results (\S\ref{sec:results}) is that it might reflect Google Street View's well-known coverage and image-quality skew towards developed regions, rather than genuine model behaviour. To rule this out, we compute three per-panorama metrics across all 1,200 evaluation samples: capture date (image age in years relative to collection date), Google's internal quality score, and empirical sharpness, defined as the variance of the Laplacian over the rendered viewport. Table~\ref{tab:data_quality} reports medians for developed and developing samples with two-sided Mann--Whitney $U$ tests; Figure~\ref{fig:coverage_confound} shows the full per-tier violin distributions.

\begin{table*}
\centering
\scriptsize
\setlength{\tabcolsep}{5pt}
\renewcommand{\arraystretch}{1.2}
\begin{tabular}{lcccc}
\toprule
\textbf{Metric} & \textbf{Developed} & \textbf{Developing} & $\boldsymbol{\Delta}$ & \textbf{$p$} \\
\midrule
Image age (years)            & 3.09  & 2.51  & $-0.58$~~(dev'g newer)   & $0.034$ \\
Quality score                & 0.979 & 0.982 & $+0.003$~~(dev'g higher) & $0.030$ \\
Sharpness (Var. Laplacian)   & 1{,}499 & 1{,}540 & $+42$~~(n.s.)        & $0.70$ \\
\bottomrule
\end{tabular}
\caption{Per-panorama image quality medians across developed and developing samples (Mann--Whitney $U$). $\Delta$ is reported as developing minus developed.}
\label{tab:data_quality}
\end{table*}

Contrary to the assumption that Street View visually favors developed regions, developing-region panoramas in our dataset are statistically \emph{newer} (median 2.51 vs 3.09 years; $p=0.034$), marginally \emph{higher} in quality score ($p=0.030$), and indistinguishable in sharpness ($p=0.70$). The per-tier violin distributions in Figure~\ref{fig:coverage_confound} further show no monotonic degradation from Tier 1 to Tier 5 in any of the three metrics. Our dataset construction (\S\ref{sec:task-setup}) had already dropped target coordinates without navigable Street View coverage, so every retained sample carries at least the coverage required for the 8-action budget. The developed/developing performance gap is therefore attributable to model behavior rather than to data availability or image-quality artifacts.

\section{Why Geo-R1 Degrades Under Navigation}
\label{app:geor1-forensics}

Geo-R1 7B is the only model in our evaluation whose agentic performance falls below its own static baselines (2549.0 km agentic vs.\ 2350.1 / 2154.4 km on 4-view / 8-view (Table~\ref{tab:model_comparison_percent}). Because Geo-R1 is explicitly fine-tuned for geospatial reasoning \citep{xu2025geor1unlockingvlmgeospatial}, we audited all 28{,}258 emitted actions across its agentic runs.

Table~\ref{tab:geor1_forensics} shows that the failure is entirely in tool calling as Geo-R1 emits an action string that the environment cannot parse in \textbf{87.1\%} of its turns. Instead of selecting one action, the model copies the pipe-delimited menu of legal actions from the prompt verbatim into the \texttt{action} field, returning strings such as \texttt{ROTATE\_LEFT|ROTATE\_RIGHT| MOVE\_FORWARD|RETURN\_START|GUESS}. The consequence is that Geo-R1 never actually navigates as it only executed 5 \texttt{MOVE\_FORWARD} actions, 1 \texttt{ROTATE\_LEFT}, and 43 \texttt{RETURN\_START} actions in total across 3{,}600 rounds. This means \textbf{99.8\%} of rounds contain no effective navigation at all.

The model's ``agentic'' condition is therefore, in substance, a single-view static evaluation conducted from the spawn heading, which is strictly less visual information than either the 4-view or the 8-view baseline receives. This explains why its agentic score lands between its random-walk and static scores.

\begin{table}[t]
\centering
\scriptsize
\setlength{\tabcolsep}{4pt}
\begin{tabular}{lcc}
\toprule
\textbf{Model} & \textbf{Invalid actions (\%)} & \textbf{Rounds w/ no nav.\ (\%)} \\
\midrule
Claude Haiku 4.5 & 0.04 & -- \\
GPT-5 Mini       & 0.00 & -- \\
Gemini 3.0 Flash & 0.00 & -- \\
Gemma 3 27B      & 0.00 & -- \\
Llama 4 Scout    & 0.00 & -- \\
\midrule
Geo-R1 7B        & \textbf{87.1} & \textbf{99.8} \\
\bottomrule
\end{tabular}
\caption{Tool-call validity under agentic navigation, over all emitted actions in three runs per model. Every general-purpose VLM produces well-formed tool calls essentially always, while the specialized Geo-R1 7B model does not.}
\label{tab:geor1_forensics}
\end{table}


Moreover, a regression with standard errors clustered by each city in a population tier (Table~\ref{tab:cluster_bootstrap}) shows that the development coefficient is not significant by any model (all $p \geq 0.13$). Therefore, the development gap is largely based on which tiers of cities are represented, which directly motivates our tier-based segregation choice beyond the developed/developing division.

\begin{table*}[ht]
\centering
\scriptsize
\begin{tabular}{lccccc}
\hline
\textbf{Model} &
\textbf{$\Delta$ mean dist (km)} &
\textbf{$\Delta$ city acc} &
\textbf{$\Delta$ country acc} &
\textbf{Dev. coef. log-dist OLS} &
\textbf{Dev. coef. city logit} \\
& \textbf{[95\% CI]} & \textbf{[95\% CI]} & \textbf{[95\% CI]} & \textbf{(p)} & \textbf{(p)} \\
\hline
GPT-5 Mini
& $-792\ [-1261,-392]$
& $-0.022\ [-.155,+.106]$
& $+0.071\ [-.009,+.157]$
& $-0.24\ (.46)$
& $+0.41\ (.14)$ \\

Gemini 3.0 Flash
& $-627\ [-1134,-201]$
& $-0.053\ [-.168,+.071]$
& $+0.025\ [-.027,+.084]$
& $-0.37\ (.14)$
& $+0.32\ (.13)$ \\

Claude Haiku 4.5
& $-1057\ [-1891,-252]$
& $+0.046\ [-.036,+.134]$
& $+0.070\ [-.056,+.196]$
& $+0.30\ (.25)$
& $-0.16\ (.59)$ \\

Gemma 3 27B
& $-1045\ [-1649,-485]$
& $-0.007\ [-.108,+.089]$
& $+0.041\ [-.069,+.151]$
& $-0.00\ (1.00)$
& $+0.19\ (.44)$ \\

Llama 4 Scout
& $-864\ [-1699,-126]$
& $+0.017\ [-.057,+.091]$
& $+0.048\ [-.055,+.158]$
& $-0.06\ (.82)$
& $+0.33\ (.22)$ \\

Geo-R1
& $\mathbf{-1194}\ [-1913,-544]$
& $+0.030\ [-.107,+.166]$
& $+0.074\ [-.045,+.194]$
& $+0.14\ (.74)$
& $+0.08\ (.83)$ \\
\hline
\end{tabular}
\caption{Cluster bootstrap results by developed minus developing cities on the agentic condition over 10 iterations. Negative distance delta = better in developed cities.}
\label{tab:cluster_bootstrap}
\end{table*}

\section{Subregion Breakdown and Retention}
\label{app:subregion}

We constructed the GeoAgent corpus through a two-step filtering process. First, we sampled a 10{,}000-row reference collection using the tier design in Table~\ref{tab:tier_dataset_composition}, which overweigh small and remote cities. Table~\ref{tab:subregion_retention} shows the characteristics of this collection in the \textbf{\% ref.} column. Noticing significant variability in the reference corpus, we sub-sampled the corpus across 18 of the 22 UN M49 subregions to ensure balanced representation and fit our design requirement. This gave us a 1{,}994-row candidate pool, which was further stratified to the final 1{,}200-row evaluation set after filtering for unnavigable coordinates. No region was significantly over- or under-represented in this process (every subregion shift is at most 0.9 percentage points), helping us construct a corpus true to the designated tier stratification.

\begin{table}[t]
\centering
\scriptsize
\setlength{\tabcolsep}{3pt}
\begin{tabular}{lrrrr}
\toprule
\textbf{UN M49 Subregion} & \textbf{$n$} & \textbf{\% ref.} & \textbf{\% parent} & \textbf{\% sample} \\
\midrule
Northern America        & 294 & 22.8 & 24.0 & 24.5 \\
Northern Europe         & 139 & 11.2 & 11.3 & 11.6 \\
South America           & 136 &  3.5 & 11.0 & 11.3 \\
Southern Asia           & 124 &  3.8 & 10.6 & 10.3 \\
South-Eastern Asia      & 101 &  5.8 &  8.0 &  8.4 \\
Eastern Europe          &  77 &  6.0 &  5.7 &  6.4 \\
Eastern Asia            &  59 &  5.2 &  4.9 &  4.9 \\
Southern Europe         &  55 &  6.0 &  4.4 &  4.6 \\
Western Europe          &  42 & 14.5 &  3.9 &  3.5 \\
Australia \& N.\ Zealand &  42 &  7.5 &  3.5 &  3.5 \\
Eastern Africa          &  30 &  1.6 &  2.2 &  2.5 \\
Southern Africa         &  29 &  1.7 &  3.3 &  2.4 \\
Western Africa          &  25 &  1.8 &  3.0 &  2.1 \\
Central America         &  14 &  6.2 &  1.2 &  1.2 \\
Western Asia            &  14 &  1.5 &  1.8 &  1.2 \\
Melanesia               &   8 &  0.7 &  0.6 &  0.7 \\
Northern Africa         &   6 &  0.8 &  0.5 &  0.5 \\
Polynesia               &   5 &  0.7 &  0.4 &  0.4 \\
\midrule
\textbf{Total}          & \textbf{1{,}200} & \textbf{100} & \textbf{100} & \textbf{100} \\
\bottomrule
\end{tabular}
\caption{Composition and retention by UN M49 subregion. \textbf{\% ref.} is an independently sampled 10{,}000-row Street View reference collection built with the same city-targeting heuristic; \textbf{\% parent} is the 1{,}994-row candidate pool retrieved for this study; \textbf{\% sample} is the final evaluation set. All parent $\to$ sample shifts are $\leq 0.9$ pp.}
\label{tab:subregion_retention}
\end{table}

\begin{table*}[ht]
\centering
\scriptsize
\begin{tabular}{lrrrrrr}
\hline
\textbf{Sub-region (M49)} & \textbf{GPT-5 Mini} & \textbf{Gemini Flash} & \textbf{Claude Haiku} & \textbf{Gemma 3 27B} & \textbf{Llama4 Scout} & \textbf{Geo-R1 7B} \\
\hline
Western Asia      & 215   & 343   & 1,245 & 419   & 1,558 & 728   \\
Eastern Europe    & 350   & 698   & 834   & 540   & 1,387 & 300   \\
Western Europe    & 572   & 560   & 407   & 1,117 & 918   & 404   \\
Northern America  & 656   & 783   & 2,061 & 1,143 & 1,370 & 872   \\
Southern Asia     & 1,173 & 1,202 & 1,507 & 1,263 & 1,279 & 1,159 \\
South America     & 1,326 & 1,151 & 3,810 & 2,841 & 4,077 & 3,483 \\
Northern Africa   & 3,961 & 2,746 & 2,231 & 2,992 & 3,615 & 4     \\
Polynesia         & 8,578 & 6,834 & 11,782 & 8,687 & 11,954 & 235   \\
Melanesia         & 11,523 & 11,203 & 11,142 & 11,797 & 11,706 & 10,524 \\
\hline
\end{tabular}
\caption{Mean Haversine distance (km) in the GeoAgent task by a few exemplar UN M49 sub-regions under the limited agentic condition.}
\label{tab:haversine_distance}
\end{table*}

We analyzed the outputs by UN M49 sub-region breakdown, with a few exemplar sub-regions in Table~\ref{tab:haversine_distance}. We find that every model collapses in Melanesia/Polynesia (mean error 10,500-11,950 km with $\leq 14\%$ country accuracy, while Northern/Western Europe and Northern America stay under ~1,100 km for the stronger models). 

Further, we ran a city-level cluster bootstrap in Table 2 stratified by development status, showing that the developed-developing mean-distance gap is significant for all six models (e.g., GPT-5 Mini $-792$ km, 95\% CI $[-1261, -392]$). However, city and country accuracies are not significant.

\section{Per-Annotator Breakdown of Human Baselines}
\label{app:per-annotator}

\begin{figure*}[htbp]
    \centering
    \includegraphics[width=0.8\textwidth]{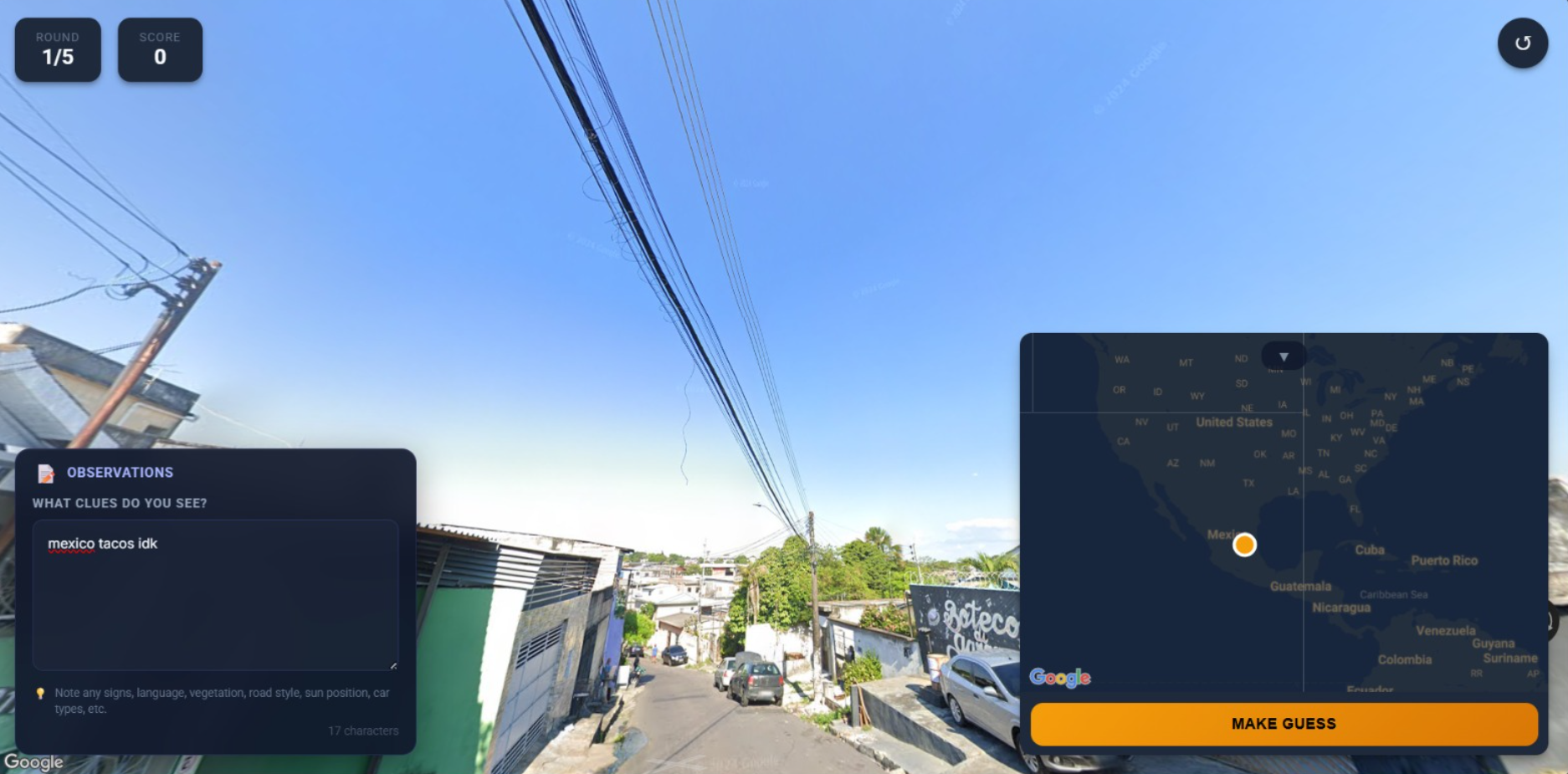}
    \caption{A screenshot of the portal for collecting human baselines. It's roughly the same as LLM agents, with the added field for adding observation chains in plain text and speech-to-text.}
    \label{fig:human_baseline}
\end{figure*}

\begin{figure}[t]
    \centering
    \includegraphics[width=\columnwidth]{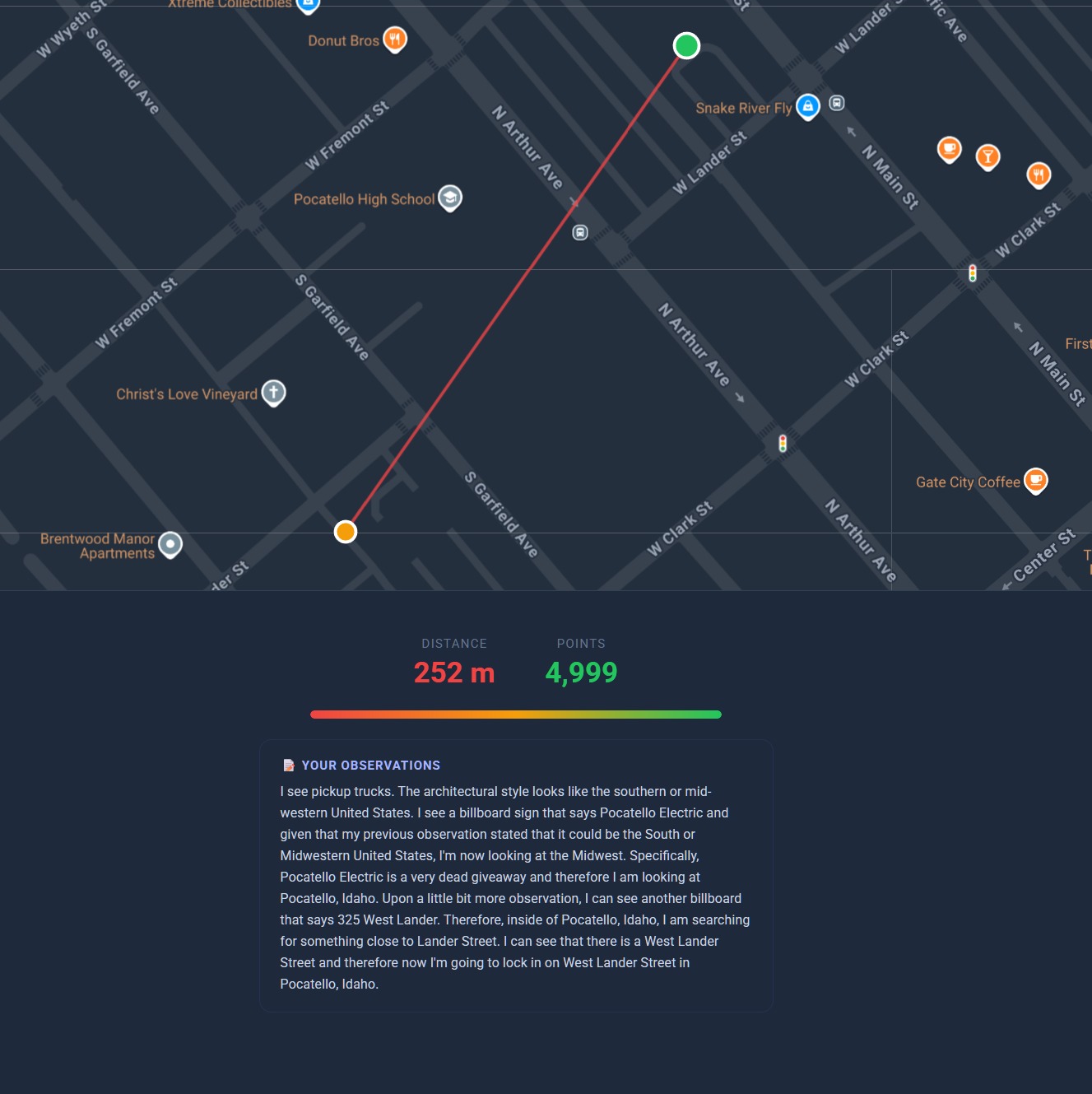}
    \caption{The human annotation pipeline allows us to collect detailed reasoning chains as we ask our participants to think out loud.}
    \label{fig:human_baseline_ann}
\end{figure}

Figures~\ref{fig:human_baseline} and \ref{fig:human_baseline_ann} show our human annotation collection software, which includes a speech-to-text feature. We require all participants to think out loud to record accurate human geolocalization reasoning chains. All annotators were recruited from KIIT University's undergraduate program and assessed through a detailed test on geographic and geolocalization performance. We selected across the spectrum, from beginner to experienced GeoGuessr capabilities, to maintain fairness in our test sample of annotators. At the end of the task, all annotators were awarded lunch coupons as payment.

Table~\ref{tab:annotator_metrics} disaggregates the human baseline reported in Table~\ref{tab:model_comparison_percent} across the six volunteer annotators. We recruited annotators with a mix of beginner and intermediate GeoGuessr exposure (no expert-level players), and each annotator scored a stratified subset drawn from the same 1{,}200-sample pool used for the model evaluations; sample sizes per annotator are reflected in the spread of the standard errors and 95\% bootstrap intervals.

First, we note that between-annotator variance is large: distance error ranges from 1{,}720 km (Annotator C) to 3{,}751 km (Annotator E), and continent accuracy from 0.20 (Annotators A and B) to 0.71 (Annotator C). Second, even the strongest human in our pool (Annotator C: 1{,}720 km mean distance, 46.8\% country accuracy, 4.3\% city accuracy) is decisively beaten by GPT-5 Mini and Gemini 3.0 Flash under both agentic and static conditions in Table~\ref{tab:model_comparison_percent}. Only Claude Haiku 4.5 under random walk approaches non-expert human performance. Third, the country-to-city accuracy gap visible across all annotators (46.8\% $\to$ 4.3\% for Annotator C; 38.0\% $\to$ 10.0\% for Annotator F) mirrors the ``last-mile'' localization failure we report for VLMs in \S4, which is an expected difficulty asymmetry of the task itself.

\begin{table*}[t]
\centering
\scriptsize
\begin{tabular}{lcccc}
\toprule
\textbf{Annotator} & \textbf{Distance Error (km)} & \textbf{Continent Acc.} & \textbf{Country Acc.} & \textbf{City Acc.} \\
\midrule
Annotator A & $2346.8 \pm 1128.0$ \; [956, 4692] & $0.200 \pm 0.133$ & $0.000 \pm 0.000$ & $0.000 \pm 0.000$ \\

Annotator B & $3050.9 \pm 1199.4$ \; [1062, 5545] & $0.200 \pm 0.133$ & $0.100 \pm 0.100$ & $0.000 \pm 0.000$ \\

Annotator C & $1720.2 \pm 270.1$ \; [1237, 2273] & $0.709 \pm 0.038$ & $0.468 \pm 0.042$ & $0.043 \pm 0.017$ \\

Annotator D & $3009.5 \pm 629.1$ \; [1829, 4319] & $0.333 \pm 0.065$ & $0.241 \pm 0.059$ & $0.111 \pm 0.043$ \\

Annotator E & $3751.4 \pm 414.7$ \; [2966, 4598] & $0.565 \pm 0.046$ & $0.348 \pm 0.045$ & $0.087 \pm 0.026$ \\

Annotator F & $3118.4 \pm 468.2$ \; [2257, 4033] & $0.640 \pm 0.069$ & $0.380 \pm 0.069$ & $0.100 \pm 0.043$ \\
\bottomrule
\end{tabular}
\caption{Per-annotator geolocation performance metrics reported as mean $\pm$ standard error, along with 95\% bootstrap confidence intervals.}
\label{tab:annotator_metrics}
\end{table*}

\section{Full TF-IDF Tables}
\label{app:tf-idf-tables}

Table~\ref{tab:country_descriptors} summarises the top-5 TF-IDF terms per country across all five evaluated models for a curated set of countries spanning developed and developing regions, and Tables~\ref{tab:tfidf_claude_agent}--\ref{tab:tfidf_openai_agent} give unabridged top-10 lists for the three agentic conditions on which we collected sufficient round-level reasoning text (Claude Haiku 4.5, Llama 4 Scout via Groq, and GPT-5 Mini). The corpora are built per the preprocessing pipeline described in \S3.2: per-round \texttt{observations} and \texttt{reasoning} fields are concatenated by ground-truth country and treated as one document per country, so the TF-IDF scores reflect what each model \emph{characteristically} writes about a region rather than its raw token frequencies.

Three patterns emerge. (i) \textbf{Granularity tracks accuracy.} GPT-5 Mini, the strongest country-level model, anchors its descriptors at city and sub-city granularity (\textit{paris, central paris, haussmann}; \textit{tokyo, kei, utility}; \textit{bangkok, soi, stayed bridge}), whereas weaker models default to script-family or regional fallbacks (\textit{north american, suburban, houses} for Llama on the US; \textit{american, flat, midwest} for Gemma on Canada). (ii) \textbf{Script recognition is near-universal but does not entail country disambiguation.} Every model surfaces \textit{bengali script}, \textit{devanagari}, \textit{thai script}, and \textit{chinese characters} for the appropriate countries, yet the same script cues trigger misattribution elsewhere. Llama 4 Scout's Switzerland entry contains \textit{chinese, chinese characters, presence chinese}, and Claude conflates Taiwan with Japan (\textit{japanese, chinese, urban, taiwan}). (iii) \textbf{Failure modes are model-specific and interpretable.} Claude Haiku 4.5's Egypt descriptors (\textit{orthodox, georgia, caucasus, tbilisi, georgian}) reveal a systematic Georgia/Egypt confusion that is invisible in the aggregate accuracy numbers; Gemma 3 27B's China entry (\textit{japanese, garden, lanterns, japanese writing}) shows a Japan-prior leak for East Asian imagery; and Llama 4 Scout's China entry contains \textit{geographic features, features signs, google logo}, indicating reliance on Street View interface chrome rather than scene content. These per-model error signatures complement the quantitative bias analysis in Figure~\ref{fig:improvement_comparison} by exposing the linguistic basis of where each model's geographic prior is brittle.

\begin{table*}[t]
\centering
\scriptsize
\setlength{\tabcolsep}{4pt}
\renewcommand{\arraystretch}{1.2}
\begin{tabularx}{\textwidth}{l X X X X X}
\toprule
\textbf{Country} &
\textbf{Claude Haiku 4.5} &
\textbf{Gemini 3.0 Flash} &
\textbf{Gemma 3 27B} &
\textbf{Llama 4 Scout} &
\textbf{GPT-5 mini} \\
\midrule
India &
indian, india, auto, tropical, market &
india, delhi, indian, rickshaws, auto &
india, hindi, indian, hindi signage, auto rickshaws &
india, tuk, english, urban, script &
indian, india, tropical, south asian, south \\
Japan &
japanese, japanese residential, ginkgo, japan, tokyo &
japanese, japan, tokyo, utility, nakano &
japanese, autumn, japanese writing, retaining, concrete retaining &
japanese, plaque, japanese characters, presence japanese, japan &
ginkgo, ginkgo leaves, japanese, leaves, tokyo \\
Norway &
nordic, norway, norwegian, road, scandinavian &
norway, norwegian, oslo, sweden, birch &
norway, scandinavian, road, sweden, norwegian &
picket, european, white picket, picket fence, white &
norway, norwegian, scandinavian, picket, birch \\
Switzerland &
swiss, tunnel, alpine, european, roofs ivy &
switzerland, bern, swiss, zurich, shutters &
german, tram tracks, tram tracks, park &
shutters, chinese, european, architecture shutters, chinese characters &
swiss, park, switzerland, european, chinese \\
Taiwan &
japanese, chinese, urban, taiwan, taiwanese &
taipei, taiwan, traditional chinese, chinese, xinyi &
chinese characters, chinese, characters, taiwan, scooters &
chinese, chinese characters, characters, presence chinese, asian &
chinese, taiwan, taipei, traditional chinese, traditional \\
Thailand &
river, bridge, cable, mekong, stayed &
bangkok, soi, thai, thailand, bridge &
river, chao phraya, chao, phraya, phraya river &
river, bridge, water, stayed bridge, cable &
river, phraya, chao, chao phraya, bangkok \\
United States &
american, midwest, road, flat, north american &
midwest, residential, alaska, united states, american &
american, lawns, maintained lawns, american cars, homes &
suburban, north, north american, houses, road &
north american, american, north, english, alaska \\
\bottomrule
\end{tabularx}
\caption{Qualitative geographic descriptors produced by different vision-language models across countries. Models generally reason with nationality-specific terms. }
\label{tab:country_descriptors}
\end{table*}


\begin{table*}[t]
\centering
\scriptsize
\setlength{\tabcolsep}{4pt}
\renewcommand{\arraystretch}{1.2}
\begin{tabularx}{\textwidth}{l X}
\toprule
\textbf{Country} & \textbf{Top Words} \\
\midrule
Argentina & spanish, andean, colombian, colonial, white, argentine, argentina, altitude, high altitude, urban \\
Australia & australian, modern, australia, melbourne, red, brick, white, tropical, trees, vegetation \\
Bangladesh & bengali, tropical, india, indian, vegetation, bengali script, bangladesh, tropical vegetation, kerala, script \\
Brazil & brazilian, brazil, tropical, portuguese, vegetation, tropical vegetation, modern, urban, portuguese signage, portuguese language \\
Cambodia & cambodia, rural, tropical, vegetation, red, soil, tropical vegetation, laterite, red laterite, laos \\
Canada & canadian, north american, american, prairie, canada, rural, midwest, trees, flat, road \\
Chile & arid, spanish, andean, peru, terrain, colonial, mexican, vegetation, palm, peruvian \\
China & elevated pedestrian, highway, modern, elevated, wet, overpass, highway overpass, urban, lighting wet, lighting \\
Egypt & orthodox, pollution, chinese, egyptian, georgia, caucasus, tbilisi, georgian, religious, fence religious \\
Estonia & european, birch, rural, nordic, road, wooden, russia, birch trees, baltic, trees \\
Eswatini & rural, vegetation, soil, road, red, landscape, highland, brazil, rwanda, terrain \\
Fiji & tropical, palm, palm trees, vegetation, white, road, island, trees, rural, caribbean \\
Finland & finland, nordic, finnish, birch, forest, birch trees, sweden, road, snow, finland sweden \\
France & paris, french, haussmann, parisian, shopping, haussmann style, arcade, pedestrian shopping, modern, shopping street \\
Germany & german, european, urban, berlin, modern, germany, brick, courtyard, radiator, german city \\
India & indian, india, hindi, tropical, typical indian, devanagari, auto, urban, devanagari script, script \\
Japan & japanese, japan, japanese residential, modern, typical japanese, distinctly japanese, suburban, urban, japanese suburban, modern japanese \\
Norway & nordic, road, birch, finland, sweden, norwegian, scandinavian, white, german, norway \\
Switzerland & swiss, german, alpine, european, switzerland, modern, green shutters, green, french, tunnel \\
Taiwan & chinese, taiwan, urban, chinese characters, taiwanese, tropical, characters, taipei, modern, chinese city \\
Thailand & thai, monument, thai script, tropical, temple, thailand, bangkok, tropical vegetation, vegetation, alley \\
United Kingdom & scottish, uk, london, scotland, british, victorian, modern, white, rendered, red \\
United States & midwest, american, road, landscape, terrain, rural, trees, flat, vegetation, suburban \\
Vietnam & vietnamese, vietnam, tropical, chi minh, minh, minh city, ho chi, ho, chi, vietnamese city \\
\bottomrule
\end{tabularx}
\caption{TF-IDF qualitative geographic descriptors for Claude Haiku 4.5 under agentic navigation.}
\label{tab:tfidf_claude_agent}
\end{table*}

\begin{table*}[t]
\centering
\scriptsize
\setlength{\tabcolsep}{4pt}
\renewcommand{\arraystretch}{1.2}
\begin{tabularx}{\textwidth}{l X}
\toprule
\textbf{Country} & \textbf{Top Words} \\
\midrule
Argentina & spanish, spanish speaking, presence spanish, spanish text, latin american, latin, south, american, speaking country, speaking \\
Australia & australian, english, palm, palm trees, australia, modern, city, skyline, city skyline, cars \\
Bangladesh & bengali, bangladesh, script, bengali script, tuk, tropical, rickshaws, presence bengali, south asian, palm \\
Brazil & portuguese, portuguese speaking, brazil, presence portuguese, cars, portuguese text, latin american, speaking, speaking country, spanish \\
Cambodia & dirt, khmer, tropical, cambodia, dirt road, khmer script, southeast, shows dirt, wall, dirt path \\
Canada & north, north american, typical north, american, suburban, canada, suggest north, houses, vehicles, cars \\
China & walkway, walkway overpass, overpass, wet, chinese, characters, china, chinese characters, city, metal railing \\
Egypt & arabic, arabic script, desert, desert like, like environment, script, presence arabic, middle eastern, middle, straight \\
France & french, france, european, likely france, presence french, french signage, buildings french, claire, city, modern \\
India & india, tuk, hindi, script, english, indian, devanagari, tuks, tuk tuks, likely india \\
Japan & japanese, japanese characters, characters, japan, presence japanese, likely japan, characters sign, characters signs, location japan, japan urban \\
Norway & european, norwegian, houses, nordic, norway, red, european style, suburban, rural, brick \\
Switzerland & european, german, european style, green shutters, green, shutters, european city, tram, green dome, city \\
Taiwan & chinese, chinese characters, characters, presence chinese, taiwan, scooters, asian, characters signs, east asian, chinese speaking \\
Thailand & thai, murky, river, thailand, brown murky, murky water, thai script, water, river brown, presence thai \\
United Kingdom & uk, houses, british style, british, european, semi detached, detached, brick, detached houses, suburban \\
United States & north, suburban, north american, typical north, american, houses, states, united states, united, typical \\
Vietnam & vietnamese, vietnam, motorbikes, presence vietnamese, likely vietnam, southeast, vietnamese text, motorcycles, asian, vietnamese characters \\
\bottomrule
\end{tabularx}
\caption{TF-IDF qualitative geographic descriptors for Llama 4 Scout under agentic navigation.}
\label{tab:tfidf_groq_agent}
\end{table*}

\begin{table*}[t]
\centering
\scriptsize
\setlength{\tabcolsep}{4pt}
\renewcommand{\arraystretch}{1.2}
\begin{tabularx}{\textwidth}{l X}
\toprule
\textbf{Country} & \textbf{Top Words} \\
\midrule
Argentina & spanish, argentina, argentine, buenos aires, buenos, aires, ushuaia, spanish language, uruguay, american \\
Australia & australian, melbourne, australia, sydney, laneway, australian style, english, darwin, tropical, inner \\
Bangladesh & bengali, bangladesh, bengali script, tropical, dhaka, rickshaws, rickshaw, script, cycle, india \\
Brazil & portuguese, brazilian, brazil, tropical, portuguese language, brazilian style, indicate brazil, portuguese signage, curitiba, brazilian city \\
Canada & prairie, canadian, north american, american, north, canada, saskatchewan, alberta, canadian prairie, flat prairie \\
France & paris, french, central paris, france, parisian, haussmann, haussmann style, lib, french language, le \\
Germany & german, berlin, european, germany, bahn, german language, eu, brick, beer garden, courtyard \\
India & india, indian, devanagari, hindi, tropical, south asian, rickshaw, auto, indian city, english \\
Japan & japanese, tokyo, kei, japan, residential, japanese text, utility, japanese style, typical japanese, indicate japan \\
Norway & norway, norwegian, birch, scandinavian, sweden, norway sweden, wooden, scandinavia, nordic, oslo \\
Switzerland & swiss, switzerland, alpine, european, zurich, german, swiss style, parking, french, shutters \\
Taiwan & taiwan, taipei, chinese, traditional chinese, traditional, characters, scooters, chinese characters, taiwanese, taiwan urban \\
Thailand & thai, bangkok, soi, thailand, boat, bridge, thai script, river, tropical, stayed bridge \\
United Kingdom & uk, scotland, uk style, dash, pebble dash, pebble, london, central london, scottish, georgian \\
United States & midwestern, american, north american, north, united states, english, deciduous, ranch, united, town \\
Vietnam & vietnamese, vietnam, minh, chi minh, chi, ho chi, hanoi, ho, minh city, motorbikes \\
\bottomrule
\end{tabularx}
\caption{TF-IDF qualitative geographic descriptors for GPT-5 Mini under agentic navigation.}
\label{tab:tfidf_openai_agent}
\end{table*}


\section{Detailed  Datasheet}
\label{app:datasheet}

\thispagestyle{empty}
\pagestyle{empty}

\noindent
\textbf{
This document is based on \textit{Datasheets for Datasets} by Gebru \textit{et
al.} \cite{gebruDatasheetsDatasets2020}. Please see the most updated version
\underline{\textcolor{blue}{\href{http://arxiv.org/abs/1803.09010}{here}}}.
}

\begin{mdframed}[linecolor=\sectioncolor]
\section*{\textcolor{\sectioncolor}{
    MOTIVATION
}}
\end{mdframed}

    \textcolor{\sectioncolor}{\textbf{
    For what purpose was the dataset created?
    }
    Was there a specific task in mind? Was there
    a specific gap that needed to be filled? Please provide a description.
    } \\
    The GeoAgent evaluation set was created to benchmark the geographic-reasoning and geolocation capabilities of large vision-language models (LVLMs) in a controlled, reproducible setting. Existing evaluation corpora either focus on well-navigated regions (biasing towards Europe/North America) or lack the diversity needed to diagnose performance differences across population density, development level, and world region. The dataset is described in full in §3.1 of the paper. \\
    
    \textcolor{\sectioncolor}{\textbf{
    Who created this dataset (e.g., which team, research group) and on behalf
    of which entity (e.g., company, institution, organization)?
    }
    } \\
    The dataset was created by the paper authors for the purpose of evaluating LVLMs on the task of predicting the country and city of a Street View panorama across a stratified global sample. It is intended for evaluation only, not training. \\
    
    \textcolor{\sectioncolor}{\textbf{
    What support was needed to make this dataset?
    }
    (e.g., who funded the creation of the dataset? If there is an associated
    grant, provide the name of the grantor and the grant name and number, or if
    it was supported by a company or government agency, give those details.)
    } \\
    No external funding was received for data collection. Google Street View imagery is subject to Google's Terms of Service; all images were retrieved via the Google Street View Static API under its standard terms. \\
    
    \textcolor{\sectioncolor}{\textbf{
    Any other comments?
    }} \\
    N/A \\

\begin{mdframed}[linecolor=\sectioncolor]
\section*{\textcolor{\sectioncolor}{
    COMPOSITION
}}
\end{mdframed}
    \textcolor{\sectioncolor}{\textbf{
    What do the instances that comprise the dataset represent (e.g., documents,
    photos, people, countries)?
    }
    Are there multiple types of instances (e.g., movies, users, and ratings;
    people and interactions between them; nodes and edges)? Please provide a
    description.
    } \\
    Each instance is a single Google Street View panorama anchored to a known geographic location. For each panorama, the dataset records are in Table~\ref{tab:dataset_metadata}. \\    

    \begin{table*}[t]
    \centering
    \scriptsize
    \renewcommand{\arraystretch}{1.2}
    \begin{tabular}{p{3.2cm} p{2.2cm} p{8.8cm}}
    \toprule
    \textbf{Field} & \textbf{Type} & \textbf{Description} \\
    \midrule
    
    \texttt{image\_id} & string & MD5 hash of the panorama image bytes --- unique instance identifier \\
    
    \texttt{pano\_id} & string & Google Street View panorama ID \\
    
    \texttt{image\_filename} & string & Filename within the collection archive \\
    
    \texttt{image\_path} & string & Relative path to the stored image file \\
    
    \texttt{ground\_truth\_lat} / \texttt{ground\_truth\_lng} & float & Coordinates of the Street View camera \\
    
    \texttt{actual\_lat} / \texttt{actual\_lng} & float & Coordinates of the Google-reported actual location \\
    
    \texttt{country} & string & Country name \\
    
    \texttt{continent} & string & Continent name (one of 6) \\
    
    \texttt{city} & string & City name (one of 100) \\
    
    \texttt{region} & string & UN M49 geographic subregion (see \S Collection Process) \\
    
    \texttt{population\_tier} & string & Population-density tier: \texttt{tier\_1}--\texttt{tier\_5} \\
    
    \texttt{tier} & string & Same as \texttt{population\_tier} (added for downstream compatibility) \\
    
    \texttt{development\_level} & string & \texttt{developed} or \texttt{developing} (World Bank income group) \\
    
    \texttt{date} & string & Panorama capture month (\texttt{YYYY-MM}) \\
    
    \texttt{copyright} & string & Street View imagery rights holder \\
    
    \texttt{heading} & float & Camera heading (degrees) \\
    
    \texttt{pitch} & float & Camera pitch (degrees) \\
    
    \texttt{fov} & float & Field of view (degrees) \\
    
    \texttt{collection\_timestamp} & datetime & API retrieval timestamp \\
    
    \texttt{quality\_score} & float & Model-based image quality score (higher = better) \\
    
    \texttt{batch\_source} & string & Collection batch identifier (\texttt{batch\_XXXX}) \\
    
    \midrule
    \multicolumn{3}{l}{\textbf{Total instances:} 1,200} \\
    \multicolumn{3}{l}{\textbf{Total fields:} 23} \\
    
    \bottomrule
    \end{tabular}
    \caption{Dataset metadata fields and descriptions.}
    \label{tab:dataset_metadata}
    \end{table*}

    
    \textcolor{\sectioncolor}{\textbf{
    How many instances are there in total (of each type, if appropriate)?
    }
    } \\
    The released evaluation set contains 1{,}200 panorama instances, one per row of the metadata file. Each instance corresponds to a single Street View viewpoint with all 23 fields in Table~\ref{tab:dataset_metadata}. \\
    
    \textcolor{\sectioncolor}{\textbf{
    Does the dataset contain all possible instances or is it a sample (not
    necessarily random) of instances from a larger set?
    }
    If the dataset is a sample, then what is the larger set? Is the sample
    representative of the larger set (e.g., geographic coverage)? If so, please
    describe how this representativeness was validated/verified. If it is not
    representative of the larger set, please describe why not (e.g., to cover a
    more diverse range of instances, because instances were withheld or
    unavailable).
    } \\
    The dataset is a stratified sample drawn from a much larger candidate pool of Google Street View panoramas. We began with 100 designed target cities across five population-density tiers and sampled coordinates within a 2-mile radius of each city centre. Coordinates without a retrievable navigable Street View environment were dropped, after which a tier-stratified random sample (proportions 5/10/20/27/38 \% for Tier~1 -- Tier~5; \texttt{random\_state=42}) was drawn to yield the 1{,}200-row evaluation set. The sample is intentionally \emph{not} representative of unconstrained Street View coverage, which is heavily skewed toward developed regions and major cities; our tier-stratified design over-represents small/remote cities (Tier~5 is 38\% of the sample) to support bias-auditing analyses (\S3.1). \\
    
    \textcolor{\sectioncolor}{\textbf{
    What data does each instance consist of?
    }
    “Raw” data (e.g., unprocessed text or images) or features? In either case,
    please provide a description.
    } \\
    Each instance comprises raw Street View viewport images (640$\times$640 JPEGs at \texttt{fov=90}) captured at four cardinal headings (0°, 90°, 180°, 270°), together with the 23 metadata fields listed in Table~\ref{tab:dataset_metadata}. No cropping, resizing, watermarking, or augmentation was applied. \\
    
    \textcolor{\sectioncolor}{\textbf{
    Is there a label or target associated with each instance?
    }
    If so, please provide a description.
    } \\
    Yes. Each instance carries the ground-truth latitude/longitude of the Street View camera, the country, city, continent, UN M49 subregion, population tier (Tier~1 -- Tier~5), and development level (\texttt{developed} / \texttt{developing}). These serve as targets for the geolocation prediction task and as strata for bias-auditing analyses. \\
    
    \textcolor{\sectioncolor}{\textbf{
    Is any information missing from individual instances?
    }
    If so, please provide a description, explaining why this information is
    missing (e.g., because it was unavailable). This does not include
    intentionally removed information, but might include, e.g., redacted text.
    } \\
    No fields are missing across the 1{,}200 retained samples. UN M49 subregion labels were not returned by the API and were back-filled in post-processing using a hand-curated city$\to$subregion mapping validated against the published UN M49 standard. \\
    
    \textcolor{\sectioncolor}{\textbf{
    Are relationships between individual instances made explicit (e.g., users’
    movie ratings, social network links)?
    }
    If so, please describe how these relationships are made explicit.
    } \\
    Relationships between instances are encoded via shared metadata: instances sharing the same \texttt{city}, \texttt{country}, \texttt{region}, \texttt{population\_tier}, or \texttt{development\_level} can be grouped for stratified analysis. No other within-dataset relationships (e.g., social links) exist. \\
    
    \textcolor{\sectioncolor}{\textbf{
    Are there recommended data splits (e.g., training, development/validation,
    testing)?
    }
    If so, please provide a description of these splits, explaining the
    rationale behind them.
    } \\
    No. The dataset is evaluation-only and is released as a single test set; train/dev/test partitions are intentionally not provided to discourage repurposing for fine-tuning. \\
    
    \textcolor{\sectioncolor}{\textbf{
    Are there any errors, sources of noise, or redundancies in the dataset?
    }
    If so, please provide a description.
    } \\
    Panorama age varies substantially (\texttt{date} field ranges from 2007 to 2025; median $\approx$~2022), so older imagery may not reflect current street conditions. A minority of panoramas are user-contributed rather than \copyright~Google captures, and may exhibit different framing or exposure characteristics. The \texttt{quality\_score} field is model-derived and was not human-validated. No duplicate \texttt{pano\_id} entries are present. \\
    
    \textcolor{\sectioncolor}{\textbf{
    Is the dataset self-contained, or does it link to or otherwise rely on
    external resources (e.g., websites, tweets, other datasets)?
    }
    If it links to or relies on external resources, a) are there guarantees
    that they will exist, and remain constant, over time; b) are there official
    archival versions of the complete dataset (i.e., including the external
    resources as they existed at the time the dataset was created); c) are
    there any restrictions (e.g., licenses, fees) associated with any of the
    external resources that might apply to a future user? Please provide
    descriptions of all external resources and any restrictions associated with
    them, as well as links or other access points, as appropriate.
    } \\
    The metadata is self-contained, but the raw imagery is owned by Google and licensed under the Google Street View Terms of Service. We release the \texttt{pano\_id} for every instance so that imagery can be reconstituted via the Street View Static API. (a) Google has historically maintained pano-ID stability, but does not formally guarantee perpetual availability of any specific panorama; (b) we additionally archive the as-collected image bytes under their MD5 hash (\texttt{image\_id}) for reproducibility; (c) future users must comply with Google Street View ToS, including the standard Static API quota and attribution requirements. \\
    
    \textcolor{\sectioncolor}{\textbf{
    Does the dataset contain data that might be considered confidential (e.g.,
    data that is protected by legal privilege or by doctor-patient
    confidentiality, data that includes the content of individuals’ non-public
    communications)?
    }
    If so, please provide a description.
    } \\
    No. All imagery depicts publicly accessible exterior locations captured by Google Street View; no confidential, privileged, or non-public communications are present. \\
    
    \textcolor{\sectioncolor}{\textbf{
    Does the dataset contain data that, if viewed directly, might be offensive,
    insulting, threatening, or might otherwise cause anxiety?
    }
    If so, please describe why.
    } \\
    No content known to be offensive or distressing has been identified. The imagery is street-level outdoor footage; Google's preprocessing pipeline blurs faces and licence plates prior to publication. We did not encounter content requiring redaction during inspection of randomly drawn samples. \\
    
    \textcolor{\sectioncolor}{\textbf{
    Does the dataset relate to people?
    }
    If not, you may skip the remaining questions in this section.
    } \\
    Only incidentally. Google Street View captures public spaces and may capture passers-by; all such individuals are face-blurred by Google's pipeline prior to publication. The dataset is not about people. \\

    \textcolor{\sectioncolor}{\textbf{
    Does the dataset identify any subpopulations (e.g., by age, gender)?
    }
    If so, please describe how these subpopulations are identified and
    provide a description of their respective distributions within the dataset.
    } \\
    No human subpopulations (e.g., by age or gender) are tagged. The dataset does, however, tag \emph{geographic} subpopulations -- country, continent, UN M49 subregion, population tier (Tier~1 -- Tier~5), and development level -- for bias-auditing analyses. Distributional summaries appear in Table~\ref{tab:tier_dataset_composition} and \S3.1. \\
    
    \textcolor{\sectioncolor}{\textbf{
    Is it possible to identify individuals (i.e., one or more natural persons),
    either directly or indirectly (i.e., in combination with other data) from
    the dataset?
    }
    If so, please describe how.
    } \\
    Not from the released metadata. Re-identification of any incidentally captured person is mitigated by Google's pre-publication face- and licence-plate-blurring pipeline, which applies to all panoramas in the dataset. \\
    
    \textcolor{\sectioncolor}{\textbf{
    Does the dataset contain data that might be considered sensitive in any way
    (e.g., data that reveals racial or ethnic origins, sexual orientations,
    religious beliefs, political opinions or union memberships, or locations;
    financial or health data; biometric or genetic data; forms of government
    identification, such as social security numbers; criminal history)?
    }
    If so, please provide a description.
    } \\
    The dataset contains geographic location data (latitude/longitude of public street-level views). While these are not private locations, location-bearing imagery is sensitive in aggregate because it can enable surveillance or unauthorised location inference if misused; we discuss this explicitly in the Ethical Considerations section of the main paper. No financial, health, biometric, or governmental-identifier data is present. \\
    
    \textcolor{\sectioncolor}{\textbf{
    Any other comments?
    }} \\
    N/A. \\

\begin{mdframed}[linecolor=\sectioncolor]
\section*{\textcolor{\sectioncolor}{
    COLLECTION
}}
\end{mdframed}

    \textcolor{\sectioncolor}{\textbf{
    How was the data associated with each instance acquired?
    }
    Was the data directly observable (e.g., raw text, movie ratings),
    reported by subjects (e.g., survey responses), or indirectly
    inferred/derived from other data (e.g., part-of-speech tags, model-based
    guesses for age or language)? If data was reported by subjects or
    indirectly inferred/derived from other data, was the data
    validated/verified? If so, please describe how.
    } \\
    Imagery and camera metadata are directly observable, retrieved programmatically via the Google Street View Static and Metadata APIs. Country/continent/city labels are returned by the API and were verified against the camera coordinates (\texttt{ground\_truth\_lat}, \texttt{ground\_truth\_lng}). UN M49 subregion labels are post-processing additions, derived from a hand-curated city$\to$subregion mapping validated against the published UN M49 standard. Population-tier and development-level labels are programmatically assigned from the city design list and the World Bank income-group classification, respectively. \\
    
    \textcolor{\sectioncolor}{\textbf{
    Over what timeframe was the data collected?
    }
    Does this timeframe match the creation timeframe of the data associated
    with the instances (e.g., recent crawl of old news articles)? If not,
    please describe the timeframe in which the data associated with the
    instances was created. Finally, list when the dataset was first published.
    } \\
    All API retrievals were issued in a single 23-minute window on 2025-09-01; the exact call time is recorded in each row's \texttt{collection\_timestamp}. The retrieval timeframe does not match the creation timeframe of the underlying panoramas: capture dates range from 2007 to 2025, with a median around 2022 and roughly two-thirds of panoramas captured in 2021 or later. The dataset is first published alongside this paper. \\
    
    \textcolor{\sectioncolor}{\textbf{
    What mechanisms or procedures were used to collect the data (e.g., hardware
    apparatus or sensor, manual human curation, software program, software
    API)?
    }
    How were these mechanisms or procedures validated?
    } \\
    Custom Python collection scripts using the official Google Street View Static API (\texttt{size=640x640}, \texttt{fov=90}, headings $\{0\degree, 90\degree, 180\degree, 270\degree\}$) and Metadata API. Each retrieved panorama is validated by computing an MD5 hash of the image bytes (stored as \texttt{image\_id}) and by cross-checking the API-returned \texttt{pano\_id} against the request. The hand-curated city$\to$UN M49 mapping was cross-checked against the published UN M49 standard. \\
    
    \textcolor{\sectioncolor}{\textbf{
    What was the resource cost of collecting the data?
    }
    (e.g. what were the required computational resources, and the associated
    financial costs, and energy consumption - estimate the carbon footprint.)
    } \\
    Modest. The full retrieval ran on a single workstation in well under one five hours of wall-clock time. The Google Static and Metadata APIs incur nominal per-call fees that fall within Google's free monthly quota for the volume collected. Carbon footprint is negligible compared to model inference for the experiments in the main paper (Appendix~\ref{app:cost_analysis}). \\
    
    \textcolor{\sectioncolor}{\textbf{
    If the dataset is a sample from a larger set, what was the sampling
    strategy (e.g., deterministic, probabilistic with specific sampling
    probabilities)?
    }
    } \\
    Probabilistic, with tier-stratified sampling probabilities and a fixed random seed. For each of 100 designed target cities, we randomly sampled candidate coordinates within a 2-mile radius of the city centre; coordinates without a navigable Street View environment were dropped. Within each population tier, panoramas were sampled at fixed proportions (5\,/\,10\,/\,20\,/\,27\,/\,38\,\% for Tier~1 -- Tier~5) totalling 1{,}200 instances. \texttt{random\_state=42} for reproducibility. \\
    
    \textcolor{\sectioncolor}{\textbf{
    Who was involved in the data collection process (e.g., students,
    crowdworkers, contractors) and how were they compensated (e.g., how much
    were crowdworkers paid)?
    }
    } \\
    The paper authors. No crowdworkers, contractors, or external annotators were involved in data collection; collection ran fully automatically against the Google Street View APIs. Annotators for the human baseline (Appendix~\ref{app:per-annotator}) are unpaid volunteers and were not involved in collecting the underlying panoramas. \\
    
    \textcolor{\sectioncolor}{\textbf{
    Were any ethical review processes conducted (e.g., by an institutional
    review board)?
    }
    If so, please provide a description of these review processes, including
    the outcomes, as well as a link or other access point to any supporting
    documentation.
    } \\
    No formal IRB review was conducted. The dataset consists exclusively of publicly accessible Google Street View imagery with no human-subject interaction at collection time; this is consistent with prior Street View geolocation benchmarks. Privacy and dual-use considerations are discussed in the Ethical Considerations section of the main paper. \\
    
    \textcolor{\sectioncolor}{\textbf{
    Does the dataset relate to people?
    }
    If not, you may skip the remainder of the questions in this section.
    } \\
    Not as a primary focus. Persons may appear incidentally in Street View imagery but are face-blurred by Google before publication. The remaining questions in this subsection therefore do not strictly apply, but we provide minimal answers for completeness. \\

    \textcolor{\sectioncolor}{\textbf{
    Did you collect the data from the individuals in question directly, or
    obtain it via third parties or other sources (e.g., websites)?
    }
    } \\
    Indirectly, through Google Street View. We did not interact with any individuals during data collection. \\
    
    \textcolor{\sectioncolor}{\textbf{
    Were the individuals in question notified about the data collection?
    }
    If so, please describe (or show with screenshots or other information) how
    notice was provided, and provide a link or other access point to, or
    otherwise reproduce, the exact language of the notification itself.
    } \\
    Not applicable. Any individuals incidentally captured were photographed by Google Street View's camera vehicles in public spaces, with face-blurring applied prior to publication. Notification protocols are governed by Google's data-collection policies. \\
    
    \textcolor{\sectioncolor}{\textbf{
    Did the individuals in question consent to the collection and use of their
    data?
    }
    If so, please describe (or show with screenshots or other information) how
    consent was requested and provided, and provide a link or other access
    point to, or otherwise reproduce, the exact language to which the
    individuals consented.
    } \\
    Not applicable for the same reason: we did not interact with depicted individuals. Consent and removal mechanisms are administered by Google through Street View's reporting interface. \\
    
    \textcolor{\sectioncolor}{\textbf{
    If consent was obtained, were the consenting individuals provided with a
    mechanism to revoke their consent in the future or for certain uses?
    }
     If so, please provide a description, as well as a link or other access
     point to the mechanism (if appropriate)
    } \\
    Not applicable. Google provides a public reporting mechanism for individuals who wish to have specific Street View imagery blurred or removed; we honour those upstream decisions by referencing panoramas by \texttt{pano\_id} rather than redistributing fixed bytes, so any subsequent removal at Google's end propagates to the dataset. \\
    
    \textcolor{\sectioncolor}{\textbf{
    Has an analysis of the potential impact of the dataset and its use on data
    subjects (e.g., a data protection impact analysis)been conducted?
    }
    If so, please provide a description of this analysis, including the
    outcomes, as well as a link or other access point to any supporting
    documentation.
    } \\
    Privacy and dual-use risks are discussed in the Ethical Considerations section of the main paper, including the standard concerns around surveillance applications, unauthorised location inference, and amplification of Street View's regional coverage bias. No formal DPIA was filed because the dataset contains no personally identifying information beyond what Google has already published with face- and licence-plate-blurring applied. \\
    
    \textcolor{\sectioncolor}{\textbf{
    Any other comments?
    }} \\
    N/A. \\

\begin{mdframed}[linecolor=\sectioncolor]
\section*{\textcolor{\sectioncolor}{
    PREPROCESSING / CLEANING / LABELING
}}
\end{mdframed}

    \textcolor{\sectioncolor}{\textbf{
    Was any preprocessing/cleaning/labeling of the data
    done(e.g.,discretization or bucketing, tokenization, part-of-speech
    tagging, SIFT feature extraction, removal of instances, processing of
    missing values)?
    }
    If so, please provide a description. If not, you may skip the remainder of
    the questions in this section.
    } \\
    Yes, at the level of instance selection and label assignment rather than image manipulation. (i) Candidate coordinates without a navigable Street View environment were dropped before sampling. (ii) Panoramas with insufficient image quality were excluded. (iii) Country/continent/city labels assigned by the API metadata response were verified against the camera coordinates. (iv) Population-tier and development-level labels were assigned programmatically from the city design list and the World Bank income-group classification. (v) UN M49 subregion labels were back-filled using a hand-curated city$\to$subregion mapping. No image-level operations (cropping, resizing, watermarking, augmentation, colour normalisation) were performed. \\

    \textcolor{\sectioncolor}{\textbf{
    Was the “raw” data saved in addition to the preprocessed/cleaned/labeled
    data (e.g., to support unanticipated future uses)?
    }
    If so, please provide a link or other access point to the “raw” data.
    } \\
    Yes. The full set of API-retrieved panorama bytes and the unfiltered candidate-pool metadata are retained alongside the filtered 1{,}200-row evaluation set, so the filtering pipeline can be re-run with different criteria. All artifacts are released through the GeoAgent HuggingFace repository. \\

    \textcolor{\sectioncolor}{\textbf{
    Is the software used to preprocess/clean/label the instances available?
    }
    If so, please provide a link or other access point.
    } \\
    Yes. All collection, filtering, and label-assignment scripts are released as part of the GeoAgent code repository. \\

    \textcolor{\sectioncolor}{\textbf{
    Any other comments?
    }} \\
    N/A. \\

\begin{mdframed}[linecolor=\sectioncolor]
\section*{\textcolor{\sectioncolor}{
    USES
}}
\end{mdframed}

    \textcolor{\sectioncolor}{\textbf{
    Has the dataset been used for any tasks already?
    }
    If so, please provide a description.
    } \\
    Yes -- for the experiments reported in this paper: static 4-view and 8-view multiview baselines, random-walk baseline, and the agentic-navigation evaluation across six VLMs (\S4). \\

    \textcolor{\sectioncolor}{\textbf{
    Is there a repository that links to any or all papers or systems that use the dataset?
    }
    If so, please provide a link or other access point.
    } \\
    The GeoAgent HuggingFace repository is the canonical reference for the dataset; future uses will be listed there as they appear. \\

    \textcolor{\sectioncolor}{\textbf{
    What (other) tasks could the dataset be used for?
    }
    } \\
    Cross-modal geographic-reasoning evaluation, bias auditing across world regions or development levels, qualitative analysis of VLM reasoning over geographic categories, and as a controlled evaluation harness for future embodied geo-agents (including non-VLM agents that use Street View as an environment). The metadata fields are also suitable for analyses of Street View coverage age, regional quality skew, and licensing composition. \\

    \textcolor{\sectioncolor}{\textbf{
    Is there anything about the composition of the dataset or the way it was
    collected and preprocessed/cleaned/labeled that might impact future uses?
    }
    For example, is there anything that a future user might need to know to
    avoid uses that could result in unfair treatment of individuals or groups
    (e.g., stereotyping, quality of service issues) or other undesirable harms
    (e.g., financial harms, legal risks) If so, please provide a description.
    Is there anything a future user could do to mitigate these undesirable
    harms?
    } \\
    The tier-stratified design deliberately over-represents small / remote (Tier~5) cities so absolute accuracy numbers should not be interpreted as estimates of real-world performance on a uniform population sample. Users should always report accuracy stratified by tier and by development level (\S4). The dataset inherits Street View's documented coverage and demographic biases (e.g., higher density in developed urban areas) and our quality analysis (Appendix~\ref{app:data-quality}) characterises these. Finally, downstream models built on this data should not be deployed for surveillance or for unauthorised location inference. \\

    \textcolor{\sectioncolor}{\textbf{
    Are there tasks for which the dataset should not be used?
    }
    If so, please provide a description.
    } \\
    The dataset should not be used to train or fine-tune models -- its size and stratification are unsuited to training objectives, and using a tier-stratified evaluation set as a training distribution would distort downstream behaviour. It should also not be used to power production geolocation services, to make country-level performance claims that would generalise beyond the cities covered, or in any application that could identify or surveil specific individuals incidentally captured in Street View imagery. \\

    \textcolor{\sectioncolor}{\textbf{
    Any other comments?
    }} \\
    N/A. \\

\begin{mdframed}[linecolor=\sectioncolor]
\section*{\textcolor{\sectioncolor}{
    DISTRIBUTION
}}
\end{mdframed}

    \textcolor{\sectioncolor}{\textbf{
    Will the dataset be distributed to third parties outside of the entity
    (e.g., company, institution, organization) on behalf of which the dataset
    was created?
    }
    If so, please provide a description.
    } \\
    Yes. The full metadata (including \texttt{pano\_id} for image reconstitution) is released publicly through the GeoAgent code repository. Raw images are not redistributed; users must reconstitute them via the Google Street View Static API under Google's Terms of Service. \\

   \textcolor{\sectioncolor}{\textbf{
    How will the dataset will be distributed (e.g., tarball on website, API,
    GitHub)?
    }
    Does the dataset have a digital object identifier (DOI)?
    } \\
    Through the project website and code repository (\url{https://geoagent-benchmark.github.io}). \\
    
    \textcolor{\sectioncolor}{\textbf{
    When will the dataset be distributed?}
    } \\
    Yes, the dataset and full model outputs are being distributed through our HuggingFace repository. \\

    \textcolor{\sectioncolor}{\textbf{
    Will the dataset be distributed under a copyright or other intellectual
    property (IP) license, and/or under applicable terms of use (ToU)?
    }
    If so, please describe this license and/or ToU, and provide a link or other
    access point to, or otherwise reproduce, any relevant licensing terms or
    ToU, as well as any fees associated with these restrictions.
    } \\
    The dataset is released under the MIT license, permitting academic and research use. Since the underlying source materials are publicly available Street View imagery, no additional licensing restrictions apply beyond standard academic attribution requirements and Google's Street View Terms of Service for the image bytes themselves. Per-row copyright is recorded in the \texttt{copyright} field (81.8\% \texttt{\textcopyright\ Google}, 18.2\% individual contributors). \\

    \textcolor{\sectioncolor}{\textbf{
    Have any third parties imposed IP-based or other restrictions on the data
    associated with the instances?
    }
    If so, please describe these restrictions, and provide a link or other
    access point to, or otherwise reproduce, any relevant licensing terms, as
    well as any fees associated with these restrictions.
    } \\
    Google retains image rights as described above. User-contributed panoramas (18.2\% of the sample) carry per-contributor attribution recorded in the \texttt{copyright} field. No additional third-party restrictions apply. \\

    \textcolor{\sectioncolor}{\textbf{
    Do any export controls or other regulatory restrictions apply to the
    dataset or to individual instances?
    }
    If so, please describe these restrictions, and provide a link or other
    access point to, or otherwise reproduce, any supporting documentation.
    } \\
    None known. The imagery is drawn from publicly accessible Street View coverage and contains no export-controlled content. \\

    \textcolor{\sectioncolor}{\textbf{
    Any other comments?
    }} \\
    N/A. \\

\begin{mdframed}[linecolor=\sectioncolor]
\section*{\textcolor{\sectioncolor}{
    MAINTENANCE
}}
\end{mdframed}

    \textcolor{\sectioncolor}{\textbf{
    Who is supporting/hosting/maintaining the dataset?
    }
    } \\
    The dataset and the project repository will be maintained by the paper authors for the foreseeable future following publication. \\

    \textcolor{\sectioncolor}{\textbf{
    How can the owner/curator/manager of the dataset be contacted (e.g., email
    address)?
    }
    } \\
    Please contact the first-author through the provided email address for any queries. \\

    \textcolor{\sectioncolor}{\textbf{
    Is there an erratum?
    }
    If so, please provide a link or other access point.
    } \\
    No erratum at the time of release. Any future corrections will be posted as versioned patch CSVs in the project repository with a dated changelog. \\

    \textcolor{\sectioncolor}{\textbf{
    Will the dataset be updated (e.g., to correct labeling errors, add new
    instances, delete instances)?
    }
    If so, please describe how often, by whom, and how updates will be
    communicated to users (e.g., mailing list, GitHub)?
    } \\
    Corrections (label fixes, coordinate adjustments) will be issued as versioned patches; the original \texttt{batch\_2.csv} will remain accessible for reproducibility. Updates will be announced through the project repository. \\

    \textcolor{\sectioncolor}{\textbf{
    If the dataset relates to people, are there applicable limits on the
    retention of the data associated with the instances (e.g., were individuals
    in question told that their data would be retained for a fixed period of
    time and then deleted)?
    }
    If so, please describe these limits and explain how they will be enforced.
    } \\
    The dataset does not directly relate to identifiable individuals; the retention question is therefore not applicable. The image bytes are not redistributed and are accessed only via Google's API, whose retention is itself governed by Google's policies. \\

    \textcolor{\sectioncolor}{\textbf{
    Will older versions of the dataset continue to be
    supported/hosted/maintained?
    }
    If so, please describe how. If not, please describe how its obsolescence
    will be communicated to users.
    } \\
    Yes. Older versions will remain available through tagged releases in the project repository so that prior published results can be reproduced exactly. \\

    \textcolor{\sectioncolor}{\textbf{
    If others want to extend/augment/build on/contribute to the dataset, is
    there a mechanism for them to do so?
    }
    If so, please provide a description. Will these contributions be
    validated/verified? If so, please describe how. If not, why not? Is there a
    process for communicating/distributing these contributions to other users?
    If so, please provide a description.
    } \\
    Contributions (additional cities, corrected labels, supplementary metadata) can be proposed via pull requests to the project repository. Accepted contributions will be released as a new versioned batch (e.g., \texttt{batch\_3.csv}) with explicit attribution. \\

    \textcolor{\sectioncolor}{\textbf{
    Any other comments?
    }} \\
    N/A. \\

\end{document}